%% file: main.tex
\documentclass{article}

\PassOptionsToPackage{numbers,compress}{natbib}
\usepackage[preprint,eandd]{styles/neurips_2026}

\newif\ifreviewversion
\reviewversionfalse
\ifreviewversion
  \makeatletter
  \renewcommand{\paragraph}{%
    \@startsection{paragraph}{4}{\z@}%
                  {1.0ex \@plus 0.3ex \@minus 0.2ex}%
                  {-1em}%
                  {\normalfont\normalsize\bfseries}%
  }
  \makeatother
\fi

\usepackage[utf8]{inputenc}
\usepackage[T1]{fontenc}
\usepackage[dvipsnames]{xcolor}
\usepackage{hyperref}
\usepackage{etoc}
\AtBeginDocument{\etocdepthtag.toc{main}}
\usepackage{url}
\usepackage{graphicx}
\usepackage{booktabs}
\usepackage{soul}
\usepackage{colortbl}
\usepackage{bm}
\usepackage[linesnumbered, boxed, ruled]{algorithm2e}
\usepackage{algpseudocode}
\usepackage{makecell}
\usepackage{multirow}
\usepackage{pifont}
\usepackage[accsupp]{axessibility} 
\usepackage{dsfont}

\AtBeginDocument{%
\setlength{\abovedisplayskip}{3pt}
\setlength{\belowdisplayskip}{3pt}
\setlength{\abovedisplayshortskip}{-3pt}
\setlength{\belowdisplayshortskip}{2pt}
}

\input{macros}

\title{Can 4D Foundation Models Remember?}
\author{
  Guangzhao He \\
  Cornell University \\
  \texttt{gh466@cornell.edu}
  \And
  Hadar Averbuch-Elor\thanks{Denotes equal advising.} \\
  Cornell University \\
  \texttt{hadarelor@cornell.edu}
  \And
  Wei-Chiu Ma\footnotemark[1] \\
  Cornell University \\
  \texttt{wm347@cornell.edu}
}

\begin{document}

\let\savedmaketitle\maketitle
\maketitle

\input{figures/teaser}

\input{sections/0_abstract}
\input{sections/1_intro_new}

\input{sections/2_related}

\input{sections/3_metric_new}
\input{sections/4_dataset}
\input{sections/5_method}

\input{sections/6_experiments}

\input{sections/7_limitations}
\input{sections/8_conclusion}

\newpage
\bibliographystyle{plainnat}
\bibliography{ref}

\newpage
\appendix
\input{supp/supp}

\end{document}

%% file: macros.tex
\newcommand{\methodname}{\textsc{PersistBench}}

\newcommand{\newedit}[1]{#1}
\newcommand{\ie}{i.e.}
\newcommand{\eg}{e.g.}

\definecolor{colorfirst}{rgb}{.866,.945,0.831} 
\definecolor{colorsecond}{rgb}{1,0.98,0.83} 
\definecolor{colorthird}{rgb}{0.76,0.87,0.92}

\newcommand{\textfirst}[1]{\begingroup\setlength{\fboxsep}{0pt}\colorbox{colorfirst}{#1}\endgroup}
\newcommand{\secondtext}[1]{\begingroup\setlength{\fboxsep}{0pt}\colorbox{colorsecond}{#1}\endgroup}

%% file: figures/teaser.tex
\vspace{-15pt}

\begin{center}
  \includegraphics[width=\linewidth]{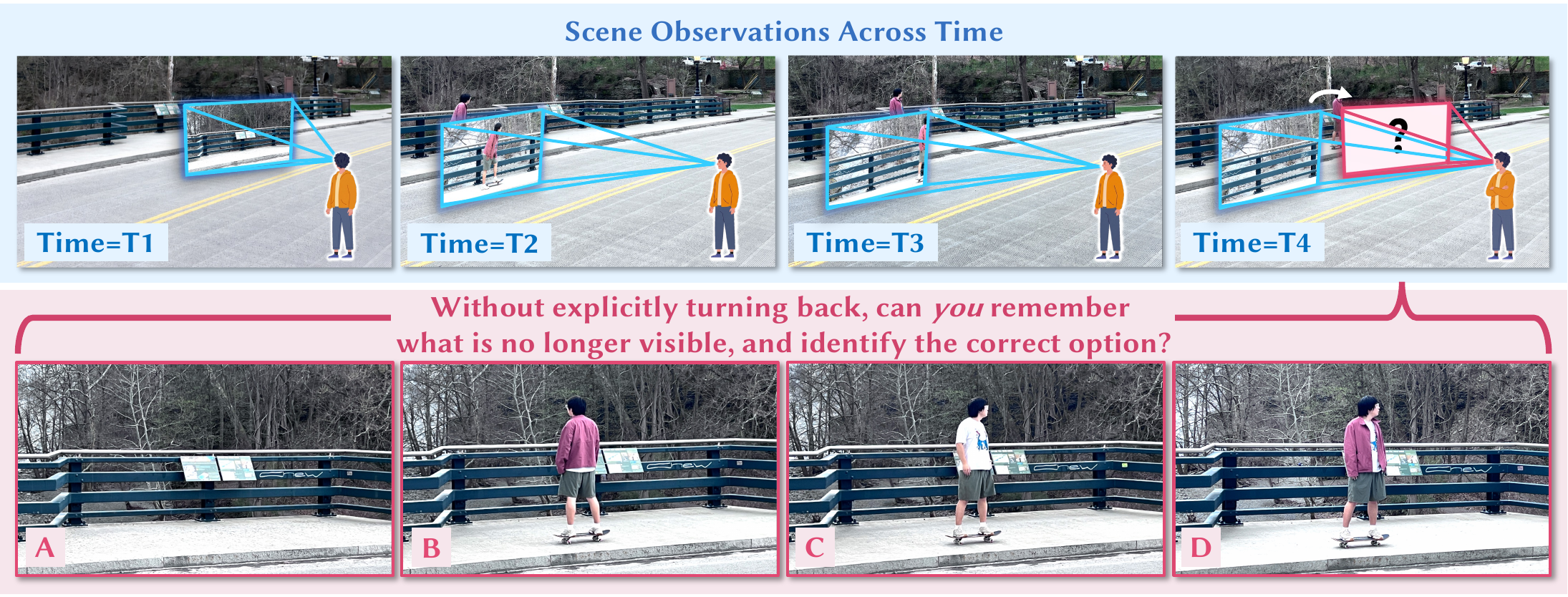}\par
  \begin{minipage}{\linewidth}
    \refstepcounter{figure}Figure~\thefigure: 
    As humans, we effortlessly maintain a dynamic mental representation of the world: after briefly observing a scene (top), we can infer what lies beyond our current field of view and anticipate how objects continue to move and appear over time. Test which option best matches \emph{your} expectation (bottom). Were you able to quickly identify that: (a) violates object permanence, (b) breaks motion continuity, and (c) fails to preserve the person's appearance?
    In this work, we explore whether 4D foundation models can remember the 4D world as we do.
    \label{fig:teaser}
  \end{minipage}
\end{center}

\vspace{8pt}

%% file: sections/0_abstract.tex
\begin{abstract}
    Perceiving and remembering the visual world is fundamental to navigating and interacting with our environment.
    Current 4D foundation models, such as camera-controllable video models or 4D reconstruction models, can perceive and reconstruct dynamic environments, but how well they remember what they have perceived remains an open question.
    Existing benchmarks largely rely on pixel-level metrics and lack ground truth for objects once they leave the field of view, making them unable to evaluate visual memory in an object-centric manner against references.
    To fill this gap, we introduce \methodname{}, a dataset and metric suite that leverages $360^\circ$ videos as omniscient ground truth and proposes three evaluation aspects: object permanence, motion continuity, and appearance preservation. 
    Evaluating various models across diverse categories reveals that current models can only maintain short-term consistency that degrades significantly once objects leave the field of view.
    Our findings highlight the gap between current model capabilities and robust visual memory ("seeing is not remembering"), providing guidance for future development of 4D foundation models. 
    Dataset and code are available on the project page: {\url{https://guangzhaohe.com/persistbench}}.
\end{abstract}

%% file: sections/1_intro_new.tex
\vspace{-1em}
\section{Introduction}
\label{sec:intro}

\begin{quote}
\vspace{-0.5em}
\small
\textit{``\dots a foolish extravagant spirit full of forms, figures, shapes, objects, ideas, apprehensions, motions, revolutions; these are begot in the ventricle of memory.''} \\
\hspace*{\fill} --- William Shakespeare, \textit{Love's Labour's Lost}
\vspace{-0.5em}
\end{quote}

Memory is a fundamental component of human cognition~\citep{visualsensory}.
It enables us to maintain and reason about the state of the world beyond immediate perception. 
Early theories, dating back to Aristotle~\cite{sorabji2006aristotle}, described memory as static impressions, akin to marks on a wax seal. 
Later perspectives in the medieval and early modern periods, reflected in the writings of William Shakespeare~\cite{shakespeare_lll_arden}, expanded this view into a richer repository of mental content that encompasses not only appearance, but also motion and temporal evolution.
Modern cognitive science studies  further refined this perspective, showing that humans develop dynamic \emph{object-centric} representations that encode how entities \emph{persist}, \emph{move}, and \emph{evolve} over time~\citep{spelke1992origins, TREISMAN198097, carey2001infants}. Such representations enable humans to reason about unobserved states of the world: for example, after  observing a person riding a skateboard, we can anticipate how they continue to move and appear even after leaving the field of view (Fig.~\ref{fig:teaser}).

This naturally raises the question: do existing \emph{4D foundation models} that reconstruct~\cite{cut3r,xu20254dgt,cognvs} or generate~\cite{gen3c,recammaster,trajcrafter} dynamic 3D scenes 
possess similar capabilities, allowing them to remember and extrapolate beyond what they observe?
Despite its apparent simplicity, answering this question is surprisingly non-trivial, for \textbf{two} critical reasons. 
\textbf{First}, there is a fundamental lack of real-world data to effectively validate out-of-view memory in dynamic scenes. 
Standard videos possess a limited field of view. When an object moves out of frame, its ongoing state is no longer captured. Without multi-viewpoint ``ground truth'' of these unobserved dynamics, we cannot quantitatively measure if a model correctly anticipates an object's evolution. 
\textbf{Second}, existing metrics predominantly focus on perceptual quality and pixel-level consistency~\citep{duan2025worldscore,huang2023vbench,zheng2025vbench2,huang2025vbench++}, falling short of assessing true memory behavior such as semantic and dynamic persistence. 
Thus, while there has been rapid progress in the development of 4D foundation models, whether these systems truly exhibit memory-like capabilities remains elusive.

With these challenges in mind, we introduce \methodname{}, 
to our knowledge, the \emph{first} benchmark that provides \emph{real-world, ground truth observations} for 
systematically evaluating visual memory in 4D foundation models. 
Our core insight is to construct real-world scenarios in which parts of the dynamic scene are unobserved to the 4D foundation models. Then we probe the models to assess whether they can accurately recover the hidden state of the world beyond its field of view.  
To overcome the data scarcity challenge, we leverage the relatively untapped resource of $360^\circ$ videos \cite{DBLP:conf/nips/WallingfordBKRD24,argus}. Unlike standard video inputs, omnidirectional videos provide full access to the environment, allowing us to define both observed inputs and withheld ground-truth reference trajectories (see Fig.~\ref{fig:360-motivation}).
Furthermore, we propose a suite of object-centric metrics that quantify three fundamental aspects of visual memory: \emph{object permanence} (e.g., does the rider in Fig.~\ref{fig:teaser} continue to exist after leaving the field of view, or is it forgotten as in option A?), \emph{motion continuity} (e.g., does the rider follow a consistent trajectory, or deviate as in option B?), and \emph{appearance preservation} (e.g., does the rider’s appearance remain consistent, or if not, to what extent does it change?). Together, these components enable a direct and principled evaluation of whether models truly maintain coherent, persistent representations of dynamic scenes.

Using \methodname, we conduct extensive evaluations of various 4D foundation models spanning multiple categories, and identify a problem that prior work has largely overlooked: current 4D foundation models lack persistent visual memory.
Our experiments show that every model performs substantially worse once the target object becomes invisible by leaving the input field of view.
This reveals a critical gap between ``seeing'' and ``remembering,'' which we attribute to a training-data bias: current 4D models are trained predominantly on videos where target objects remain visible throughout, providing little supervision for memory. 
We further analyze how different architectural design choices influence this behavior, suggesting potential directions for developing 4D foundation models with more persistent visual memory.

%% file: sections/2_related.tex
\section{Related Work}

\paragraph{4D foundation models.}
The rise of visual foundation models capable of perceiving and reasoning in 3D space~\cite{da3,flare,DBLP:conf/cvpr/Wang0CCR24,mast3r} has spurred increasing interest in models that operate over temporally-evolving 3D scenes.
This includes 4D reconstruction models, which process input video frames and reconstruct the underlying dynamic 3D scene. 
These models often use implicit representations~\citep{cut3r, chen2025ttt3r}, encoding dynamic scenes as latent sequences decoded by a neural renderer, while some prior work pursues explicit modeling~\citep{xu20254dgt, flare} that yields directly renderable representations such as Gaussian Splatting~\citep{3dgs}.
Unlike 4D reconstruction models,
4D generative models~\citep{camctrlII,recammaster,trajcrafter,gen3c,bullettime,spacetimepilot,plenopticvideogen,anyview,neoverse,cognvs,hydra}---often referred to as \emph{world models}---aim to synthesize novel views and timesteps given input observations, typically building upon video generation backbones~\citep{nvidia2026worldsimulationvideofoundation,wan2025,yang2024cogvideox,hong2022cogvideo}.
In particular, Video-to-360$^\circ$ models~\citep{argus,360anything,imagine360,viewpoint,cubecomposer} predict panoramic video from perspective inputs, and can be probed by cropping perspective views at any desired region. 
In this work, we probe the internal dynamic scene representations of these various 4D foundation models through their rendering interface.
\newedit{We leave for future work models that require tailored evaluation setups such as those that do not accept video as conditioning input~\cite{genie3,lingbotworld,matrixgame2}, those that output non-renderable 4D representations~\cite{vggt, DBLP:conf/cvpr/Wang0CCR24} (e.g., point clouds), and those that operate in the language domain~\cite{flamingo,llava} (e.g. Vision-Language Models).}

\noindent\textbf{Benchmarking visual memory.} 
Evaluating how well vision models retain and reason about what they have observed is essential for understanding their capacity to act as reliable perception systems in dynamic environments.
Vision-language model benchmarks such as VSI-Bench~\citep{yang2025thinking} and VSI-SUPER~\citep{Tong2024Cambrian1AF} probe spatial memory through question-answering about object distances and positions in egocentric videos, reducing visual memory evaluation to simple textual descriptions that fail to capture the rich visual information of the real world.
Reconstruction benchmarks~\citep{mipnerf360,Knapitsch2017,dnerf,hypernerf,dycheck,dynerf} can measure memory implicitly through photometric accuracy (PSNR, SSIM~\citep{ssim}, LPIPS~\citep{lpips}) on held-out views. 
However, by only evaluating rendering fidelity, they fail to capture the object-centric nature of visual memory.
Among generative model benchmarks~\citep{huang2023vbench,huang2025vbench++,zheng2025vbench2,DBLP:conf/nips/HuangSXLL23,DBLP:conf/nips/GhoshHS23,lin2024evaluating,DBLP:conf/cvpr/WuYLZLGLW24,duggal2025eval3d,cai2025gt23dbenchcomprehensivegeneraltextto3d}, VBench~\citep{huang2023vbench,huang2025vbench++,zheng2025vbench2}, WorldScore~\citep{duan2025worldscore}, and WCS~\citep{rakheja2025world} mainly evaluate pixel-level consistency, while concurrent efforts~\citep{ma2026sightmindevaluatingstate,duan2026liveworldsimulatingoutofsightdynamics}, perhaps most related to our setting, have begun to realize the importance of object permanence after moving out of view.
However, most of these benchmarks are fundamentally \emph{reference-free}: they evaluate visual plausibility or internal consistency without verifying whether a model faithfully retains the appearance, location, and motion state of specific observed entities. 
Our benchmark instead enables \emph{reference-based} evaluation by leveraging real-world 360$^\circ$ video as ground truth while providing only perspective crops as input, offering the first systematic evaluation of object-centric visual memory across multiple aspects with real-world data.

\noindent\textbf{Visual memory in cognitive science.}
Research in human cognition~\citep{spelke2007core, 10.1093/oso/9780190618247.001.0001} suggests that visual memory is fundamentally object-centric, composed of multiple perceptual competencies that emerge over development. Early on, humans acquire \emph{object permanence}—the ability to represent objects as continuing to exist even when they are occluded~\citep{piaget1954construction}. Violation-of-expectation studies demonstrate that even young infants exhibit surprise when objects behave inconsistently with this principle~\citep{baillargeon1987object}.
In parallel, humans develop sensitivity to \emph{spatiotemporal continuity}, expecting objects to follow coherent trajectories through space and time~\citep{SPELKE199029, spelke1992origins, TREISMAN198097}. This capability supports tracking and prediction of future states~\citep{hespos2012physics}. In contrast, \emph{appearance-based} representations emerge later: infants initially rely on spatiotemporal cues and often fail to individuate objects based on featural differences under occlusion, with robust appearance-based reasoning developing only subsequently~\citep{XU1996111, carey2001infants}.
Together, these findings indicate that visual memory is not a monolithic ability, but rather a hierarchy of object-centric representations. This motivates our approach of decomposing visual memory into distinct, measurable components for evaluating computational models.

%% file: sections/3_metric_new.tex
\input{figures/overview.tex}

\vspace{-6pt}

\section{Quantifying Visual Memory}
\vspace{-2pt}

\label{sec:quant_memory}

Although there is a common intuition that certain 4D foundation models remember better than others, no prior work has \emph{systematically} quantified this phenomenon. Many fundamental questions remain unanswered: do these models know an object still \emph{exists} after it leaves the frame?  
Do they maintain \emph{consistent motion} for moving objects beyond the camera's view?
Can they preserve the object's \emph{appearance} when revisited from a new viewpoint?

This gap in evaluating visual memory stems from two main issues. First, there is no formal definition of what it means for a model to \emph{remember}. Second, existing datasets lack ground truth for regions beyond the camera's view. 
In this section, we formally define visual memory, characterize its essential properties, and operationalize these properties into concrete computational metrics.
We address the data challenge in Sec.~\ref{sec:data_memory}.

\subsection{Problem Formulation}
\label{sec:prob-form}
Drawing on findings from cognitive science regarding how humans perceive and reason about the physical world~\citep{spelke2007core, 10.1093/oso/9780190618247.001.0001, hespos2012physics}, 
we define a model's ability to \emph{remember} as its capacity to (i) {maintain the existence of observed objects}, (ii) {retain their spatial location and motion state}, and (iii) {preserve their visual appearance} over time and across viewpoints not observed in the inputs. 

Formally, given a set of input observations $\{\mathcal{I}_i\}_{i=1}^{N}$ and a 4D foundation model $f$, we first query the model to generate outputs $\{\mathcal{O}_t\}_{t=1}^{T}$ using a set of target probes $\{\bm{p}_t\}_{t=1}^{T}$:
\begin{equation}
\{\mathcal{O}_t\} = f(\{\mathcal{I}_i\}, \{\bm{p}_t\}).
\end{equation}
We then quantify the model's memory by comparing its generated outputs against ground-truth references $\{\mathcal{O}^{\text{ref}}_t\}_{t=1}^{T}$ across various criteria $g$:
\begin{equation}
\texttt{Score} = g(\{\mathcal{O}_i\}_{i=1}^{T}, \{\mathcal{O}^{\text{ref}}_i\}_{i=1}^{T}).
\end{equation}
As we will detail in Sec.~\ref{sec:obj}, $g$ can be instantiated in several ways. For instance, it could measure how well the model preserves visual appearances.

While the inputs $\{\mathcal{I}_i\}$ and outputs $\{\mathcal{O}_t\}$ of modern 4D foundation models can take various forms, we assume both to be videos in this work, as this setup encompasses a wide range of models (see Sec.~\ref{sec:exp}). However, we note that this formulation easily extends to other modalities (e.g., point clouds). 

We make no assumptions about the model's internal architecture and interact with it solely through the target probes, which we define as camera trajectories $\{\bm{p}_t \in \mathbb{SE}(3)\}$. Intuitively, by setting these probes to novel viewpoints different from the original input, we force the model to generate scenes that could be completely invisible in the input. This allows us to validate whether the model has truly built a persistent, object-centric 4D representation or is merely memorizing 2D pixels.

\subsection{Operationalizing Visual Memory Evaluation}
\label{sec:obj}
Effectively quantifying visual memory requires carefully designed evaluation criteria $g$. Unfortunately, existing metrics focus on pixel-level accuracy~\citep{duan2025worldscore,huang2023vbench}, conflating a model's memory capability with its rendering fidelity. To decouple these concepts, we introduce three distinct yet complementary metrics designed to isolate and capture the fundamental properties of visual memory. 

\noindent\textbf{Object permanence.} 
Our first metric evaluates whether an object observed in the input sequence continues to exist within the model's underlying representation, even after leaving the original field of view. If a model possesses true object permanence, it should successfully render the object when queried with a novel viewpoint where that entity is expected to be visible.

To operationalize this, we evaluate the generated outputs $\{\mathcal{O}_t\}_{t=1}^{T}$ using a dual-verification mechanism. Given a reference mask of the target object from the input, we first utilize SAM2~\citep{sam2} to track the object across the generated frames. If a model truly ``remembers'' an object, when probed with the right viewpoint, it should consistently generate an entity with sufficient coherence to track. 

However, because a tracker might continue following a spatial region even if its visual identity severely drifts or degrades, we further utilize a VLM to evaluate appearance fidelity~\citep{zhang2023gpt4visiongeneralistevaluatorvisionlanguage,feizi2025pairbenchvisionlanguagemodelsreliable,DBLP:conf/cvpr/WuYLZLGLW24}. We consider an object to be successfully preserved in a given frame only if it satisfies both criteria: the tracker produces a valid mask, and the VLM confirms the target object's presence.

Finally, the object permanence score is defined as the percentage of valid frames across the sequence:
\begin{equation}
\texttt{Perm} = \frac{1}{T} \sum_{t=1}^{T} \mathds{1}\![\text{SAM}_t \land \text{VLM}_t],
\end{equation}
where $\text{SAM}_t \in \{0, 1\}$ and $\text{VLM}_t \in \{0, 1\}$ are binary indicators of the tracker's and VLM's response at frame $t$, and $\mathds{1}\![\cdot]$ is the indicator function.

\noindent\textbf{Motion continuity.} 
Besides permanence, a model must continuously update an entity's state based on observed dynamics, regardless of whether the object remains within the field of view or not. Static objects should remain stationary, while dynamic objects should naturally continue their motion based on prior trajectory patterns.

To assess this capability, we evaluate the spatial alignment between the predicted and ground-truth object locations over time. Building upon the tracking masks extracted during the permanence evaluation, we first compute the $\ell_2$ distance between the centroids of the predicted object mask $\mathbf{c}_t^{\text{pred}}$ and the reference object mask $\mathbf{c}_t^{\text{ref}}$ at each frame $t$. 

Because an absolute pixel distance implies different error severities depending on the object's scale, we further normalize this distance using the average area of the reference object $\overline{A}^{\text{ref}}$. Finally, following~\citep{cocooks}, we define motion continuity as an exponentially decayed, scale-normalized distance:
\begin{equation}
    \texttt{Cont} = \frac{1}{T} \sum_{t=1}^{T} \exp\!\left( -\frac{\alpha \, \|\mathbf{c}_t^{\text{pred}} - \mathbf{c}_t^{\text{ref}}\|_2}{\sqrt{ \overline{A}^\text{ref} }} \right),
\end{equation}
where $\alpha > 0$ is a temperature hyperparameter. Empirically, we set $\alpha=1.0$.
In practice, since \texttt{Cont} is designed to capture motion continuity which assumes the object exists, we compute it over frames where the object is successfully tracked (i.e., $\text{SAM}_t = 1$).
\newedit{We examine alternative motion continuity definitions and discuss the ambiguity of reference-based motion evaluation in Supp. Mat., Secs.~\ref{sec:motion-continuity-ablation} and~\ref{sec:motion-ambiguity}, respectively.}

\noindent\textbf{Appearance preservation.}
Our final metric (\texttt{App}) evaluates how well a model preserves an object's visual identity. 
Compared to object permanence, it provides a more fine-grained estimate of the output's visual faithfulness. 
We focus on measuring feature similarity~\citep{dinov2} between the generated entity and the ground-truth observation, as these features capture rich visual semantics that are highly robust to scale and lighting variations.

Specifically, at each time step $t$, we extract DINOv2~\cite{dinov2} feature maps $\mathbf{F}_t^{\text{pred}}$ and $\mathbf{F}_t^{\text{ref}}$ from the generated and reference frames. To isolate the target object, we apply the corresponding object masks and perform spatial average pooling to produce a single feature vector for each. 
Finally, we compute the appearance preservation score as the average cosine similarity between these pooled feature vectors across the generated sequence:
\begin{equation}
    \texttt{App} = \frac{1}{T} \sum_{t=1}^{T} \cos\!\left( \phi(\mathbf{F}_t^{\text{pred}}),\; \phi(\mathbf{F}_t^{\text{ref}}) \right),
\end{equation}
where $\phi(\cdot)$ denotes the masked average pooling operation. By comparing global semantic features rather than raw pixels, this metric remains invariant to minor pose shifts and partial occlusions.
Similar to \texttt{Cont}, we compute \texttt{App} over frames where the object is successfully tracked, so that the metric captures appearance fidelity regardless of how long the object exists in the model's memory.
\newedit{Ablations of the object permanence and appearance score designs, together with a comparison of tracked-frame and all-frame aggregation, are provided in Supp. Mat., Secs.~\ref{sec:permanence-score-ablation}, \ref{sec:appearance-score-ablation}, and~\ref{sec:tracking-conditioned-aggregation}.}

%% file: figures/overview.tex
\begin{figure}[t]
    \centering
    \includegraphics[width=\linewidth]{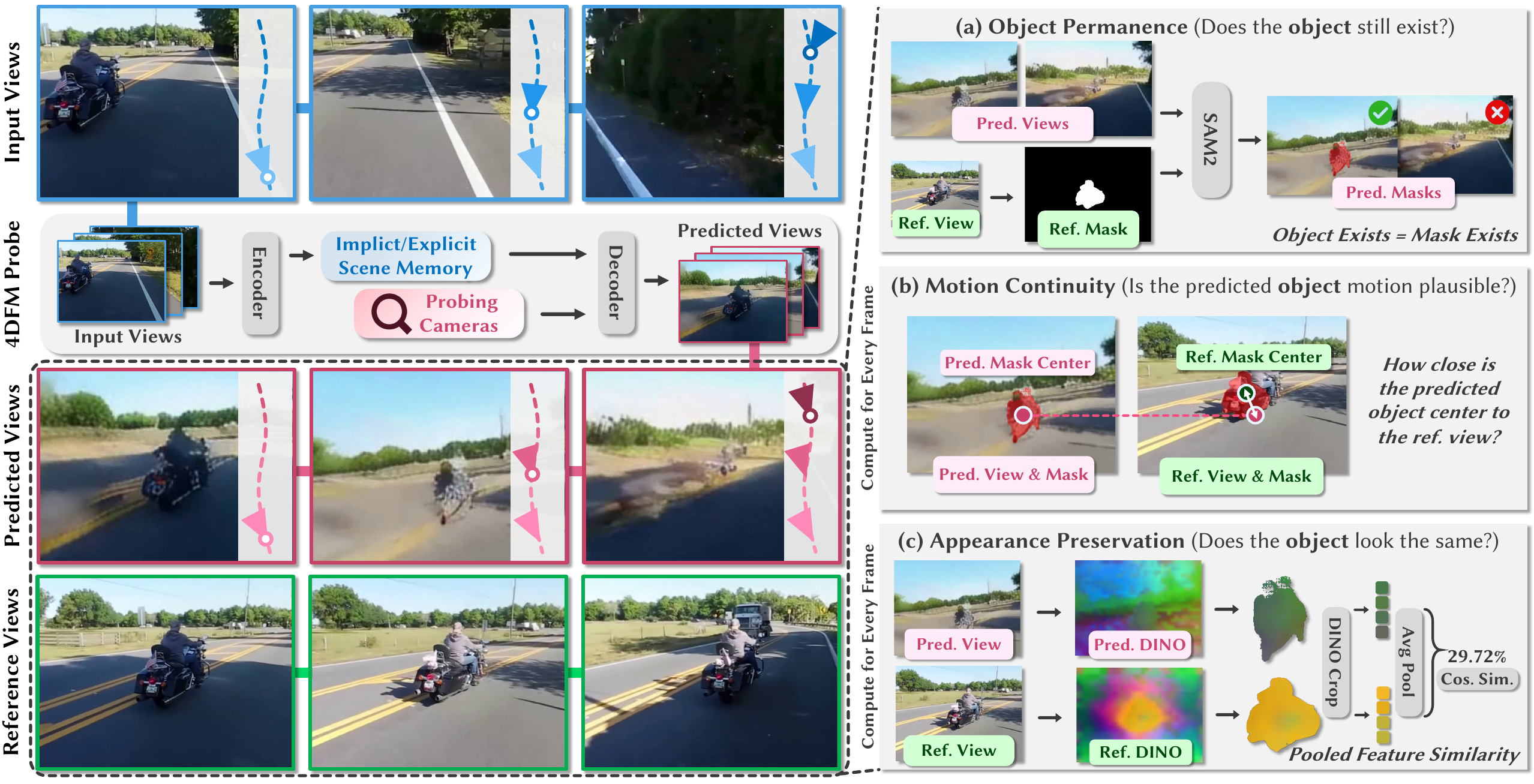}
    \vspace{-15px}
    \caption{\textbf{Evaluating visual memory.}
    We construct input videos to 4D foundation models, where an observed object (motorcycle) leaves the field of view, and then probe the models with virtual cameras to render its internal memory.
    Using the ground-truth reference views, we evaluate three object-centric aspects of visual memory, as illustrated on the right.
    }
    \label{fig:overview}
    \vspace{-10pt}
\end{figure}

%% file: sections/4_dataset.tex
\input{figures/360_motivation}

\vspace{-20pt}

\section{Evaluating Visual Memory with 360$^\circ$ Video}
\label{sec:data_memory}

\subsection{360$^\circ$ Video as Natural Visual Memory Testbed}
Now that we have established how to quantify visual memory, the remaining challenge is sourcing the right data for evaluation: \emph{on which data can we most rigorously test these 4D foundation models?}

Ideally, such data should capture not only real-world dynamics but also provide comprehensive multi-view observations. This ensures we can rigorously validate a model's predictions even after an entity exits the original field of view. Unfortunately, standard videos are not sufficient as once an object leaves the frame, its ongoing state is no longer captured.

To overcome this data challenge, we leverage the relatively untapped resource of 360$^\circ$ videos. Unlike common perspective recordings, a 360$^\circ$ video captures the full surrounding environment at every timestep (Fig.~\ref{fig:360-motivation}). It serves as an omniscient record that can be freely cropped into normal perspective views at any desired camera angle \emph{post-capture}. This flexibility allows us to construct input videos by specifying arbitrary camera trajectories, strictly controlling exactly what the model observes. Crucially, for any object that is not physically occluded, we can simply obtain its ground truth by directing a virtual camera toward its location at any point in time. This provides a reliable reference to evaluate whether the model retains its internal representation of the entity beyond direct observation. Next, we describe in detail how we leverage this boundless field of view to derive high-quality video pairs for memory evaluation.

\subsection{Extracting Input-Reference Pairs}
\label{sec:making-video-pairs}
Given a $360^\circ$ video and its camera poses, our goal is to extract a \emph{pair} of perspective video sequences: 
an \emph{input} sequence that serves as the model's input, and a \emph{reference} sequence that serves as the ground truth for evaluation. 
Both sequences are paired with their respective camera poses, which act as target probes for the generation process.

To effectively evaluate the three core properties of visual memory, we partition the input into visible and invisible segments.
Specifically, we devise a camera trajectory sampling strategy where the object of interest is initially visible but leaves the field of view in later frames of the input sequence; \newedit{distributions of visible and invisible segment durations are provided in Supp. Mat., Sec.~\ref{app:segment-duration-statistics}.} 
In contrast to the input sequence, the object remains visible throughout the reference sequence. The 4D foundation model is tasked to synthesize the reference sequence using the input sequence as its only signal of the world's state. 
If the model can render the object in the reference viewpoint accurately across both visible and invisible segments,
we say it demonstrates true persistent visual memory. 
Our dataset construction pipeline consists of two steps: 

\input{figures/example_results.tex}

\noindent\textbf{Tracking objects of interest.} 
We begin by identifying salient objects in the 360$^\circ$ video using a VLM labeler~\cite{qwen3vl}, which generates a text description for each object.
Since our evaluation requires precise control over when each object enters and exits the camera's field of view, we need to know its position at every timestep.
We therefore use SAM3~\citep{sam3} to segment and track all object instances throughout the video.
One challenge is that 360$^\circ$ videos are stored in equirectangular format, where the scene wraps around at the left and right edges.
Naively applying a tracker in this format fails when objects cross this boundary, producing discontinuous tracks that cannot associate the same object across time.
To resolve this, we horizontally duplicate the equirectangular canvas and run tracking on the augmented clip, ensuring that an object crossing one edge always has a continuous duplicate near the center.
The resulting tracks are then filtered and mapped back to the original frame to produce seamless and persistent object trajectories.

\noindent\textbf{Constrained camera trajectory sampling.} 
Given object tracks, the next step is to come up with camera trajectories for the input and reference sequences such that they are physically plausible, free of sudden jitters, and satisfy our visibility constraints (\ie, each object enters and exits the field of view at the intended times).
Since naively interpolating camera poses often results in abrupt motion or repeated pattern, we follow prior work in motion planning~\citep{chomp, stomp, trajopt} and formulate this as a trajectory optimization problem.
Specifically, we represent camera trajectories as B-splines~\citep{bsplines} and optimize the control points using L-BFGS~\citep{byrd1995limited} to balance visibility constraints with kinematic smoothness, minimizing velocity, acceleration, and jerk.
See Supp. Mat., Sec.~\ref{app:dataset} for more details.

Once we obtain the camera trajectories for both the input and reference sequences, we crop the corresponding perspective views from the 360$^\circ$ video. 
This yields paired video clips that serve as a rigorous testbed for our visual memory evaluation. An example video pair is shown in Fig.~\ref{fig:360-motivation}.

%% file: figures/360_motivation.tex
\begin{figure}[t]
  \centering
  \includegraphics[width=\linewidth]{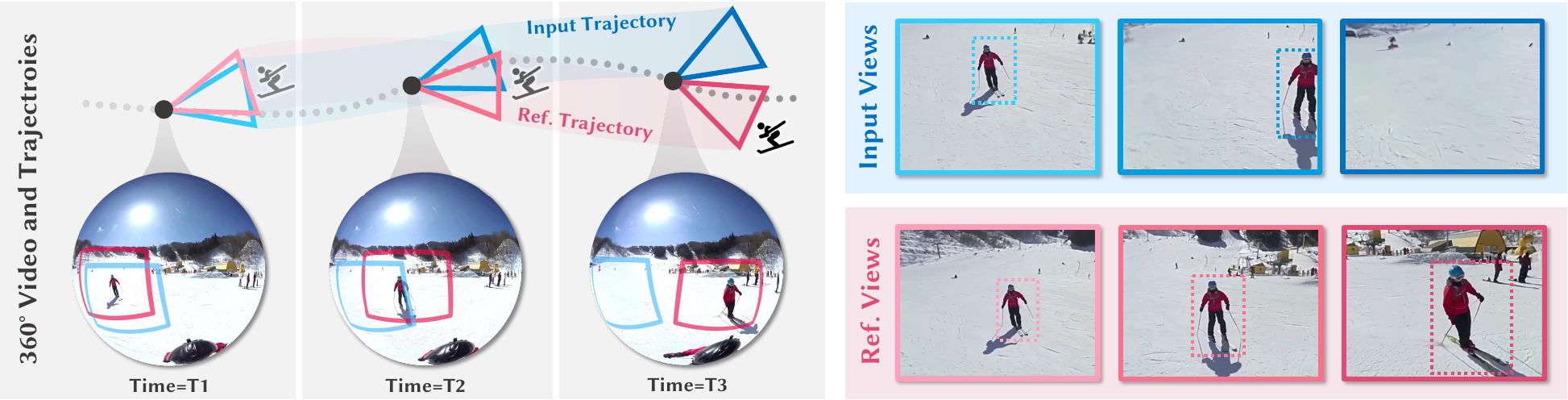}
  \vspace{-15pt}
  \caption{\textbf{Evaluating with 360$^\circ$ video.} Using 360$^\circ$ video enables arbitrary camera projections of the same underlying scene. Given one 360$^\circ$ video and a target object, we generate two camera trajectories (input and ref.). In the input trajectory, the target object is visible during the first half and leaves the view in the second. In the reference (Ref.) trajectory, the object remains in frame throughout, providing a ground-truth reference even after it has left the input view.}  
  \label{fig:360-motivation}
  \vspace{-15pt}
\end{figure}

%% file: figures/example_results.tex
\begin{figure}[t]
    \centering
    \includegraphics[width=\linewidth]{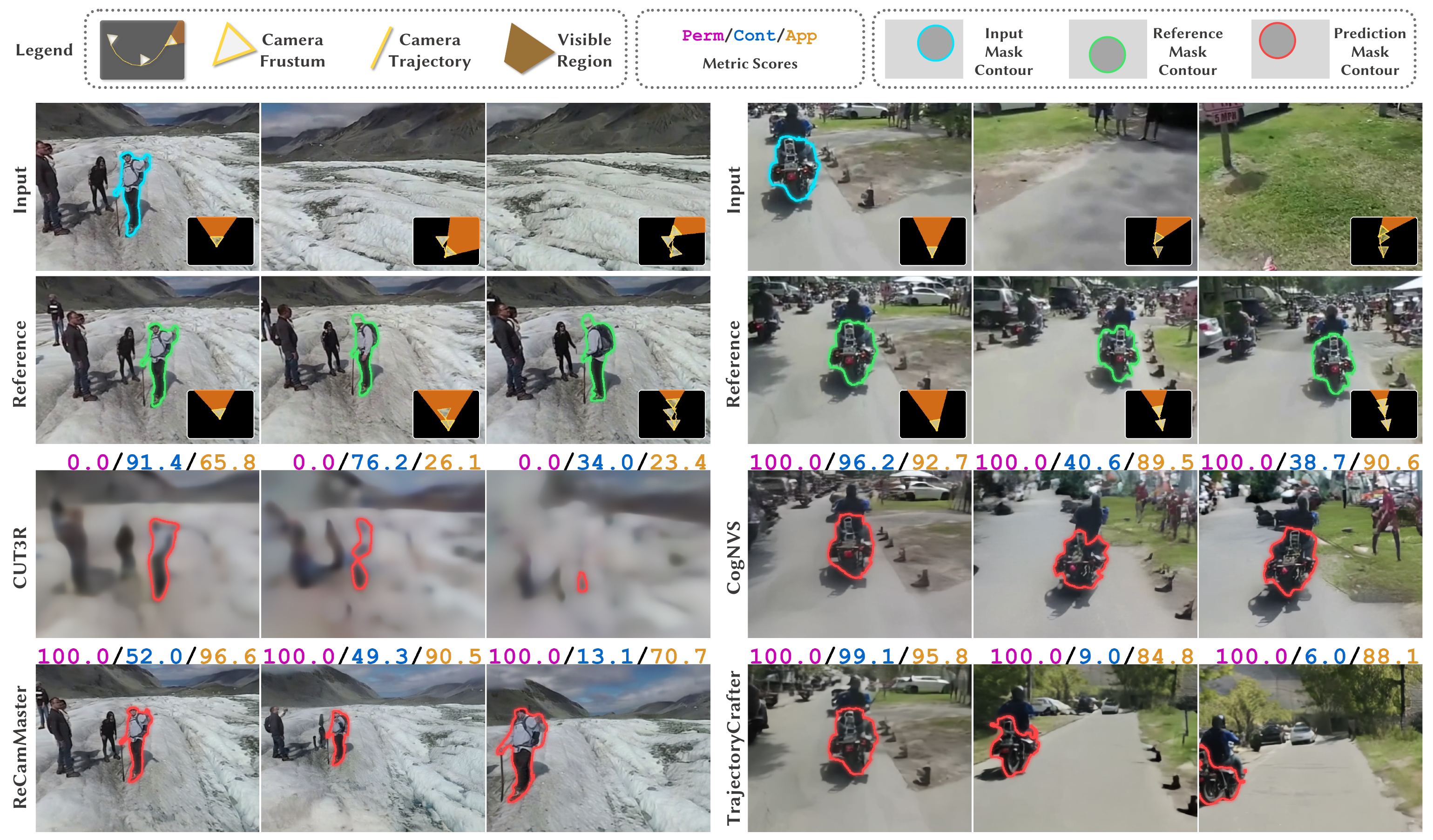}
    \vspace{-15pt}
    \caption{\newedit{
    \textbf{Qualitative comparisons with per-frame metrics.}}
    We show per-frame evaluation results for two cases, each from two representative models: CUT3R~\cite{cut3r}, ReCamMaster~\cite{recammaster}, CogNVS~\cite{cognvs} and TrajectoryCrafter~\cite{trajcrafter}.
    SAM2 tracking masks are highlighted.
    \textbf{Left (CUT3R):} our permanence score \texttt{Perm} correctly identifies cases where SAM2 maintains a valid track but the predicted object identity has degraded beyond recognition.
    \textbf{Left (ReCamMaster):} our motion continuity score \texttt{Cont} accurately reflects whether the predicted object motion agrees with the reference views.
    \textbf{Right (TrajectoryCrafter):} our appearance preservation score \texttt{App} robustly captures appearance consistency regardless of motion prediction quality.
    More results for all models are provided in Supp. Mat., Sec.~\ref{supp:full-results}.
    \newedit{We refer readers to our \href{https://guangzhaohe.com/persistbench}{project page} for more interactive visualizations.}}
    \label{fig:example-results}
    \vspace{-12pt}
  \end{figure}

%% file: sections/5_method.tex
\vspace{-8pt}
\section{The \methodname{}}
\vspace{-5pt}

Our final step is to leverage our carefully designed metrics (Sec. \ref{sec:quant_memory}) and our data curation pipeline (Sec. \ref{sec:data_memory}) to construct a rigorous evaluation suite (dubbed \methodname{}) for testing visual memory.

\label{sec:dataset}
We construct \methodname{} from a subset of the publicly available 360-1M~\citep{DBLP:conf/nips/WallingfordBKRD24} dataset. We first exploit a SLAM system~\citep{vipe} to compute initial camera poses, and then leverage GeoCalib~\citep{geocalib} to align camera orientations with gravity. 
We discard any $360^\circ$ clips where pose estimation fails or where the subsequent object tracking and B-spline trajectory optimization do not meet our smoothness and visibility constraints. From an initial pool of 24K raw $360^\circ$ clips, this process yields 2,000 high-quality, paired evaluation sequences. 
To analyze model performance across different object types, we use a VLM \citep{qwen3vl} to categorize clips into 10 categories, including human, vehicle, animal, structure, and furniture.
The clips are further divided into static and dynamic subsets based on whether the target object moves within the scene.
These pairs represent a diverse range of complex motion and appearance scenarios, providing a statistically significant foundation for evaluating the limits of visual memory in foundation models.

\newedit{Our automated evaluation pipeline uses off-the-shelf models: DINOv3~\citep{dinov3} localizes the target object in the first visible generated frame, SAM2~\citep{sam2} tracks it through subsequent frames, an open-weight VLM Qwen3.8-27B~\citep{qwen38} serves as the VLM judge to verify object existence, and DINOv2~\citep{dinov2} provides features for the \texttt{App} score.}

%% file: sections/6_experiments.tex
\input{tables/main_table_v2}

\vspace{-5pt}
\section{Results}
\vspace{-5pt}

\label{sec:exp}

We evaluate 12 publicly available 4D foundation models spanning three distinct categories: 
(i) \textbf{4D reconstruction models} supporting dynamic novel-view synthesis (CUT3R~\cite{cut3r}, 4DGT~\cite{xu20254dgt}, CogNVS~\cite{cognvs}, NeoVerse~\cite{neoverse}); 
(ii) \textbf{camera-controlled video models} (ReCamMaster~\cite{recammaster}, TrajectoryCrafter~\cite{trajcrafter}, GEN3C~\cite{gen3c}, HyDRA~\cite{hydra}); 
and (iii) \textbf{video-to-360$^\circ$ generation models} (ViewPoint~\cite{viewpoint}, Imagine360~\cite{imagine360}, Argus~\cite{argus}, CubeComposer~\cite{cubecomposer}), 
which we evaluate by cropping the generated view to ensure a fair perspective comparison.

We evaluate all models across both visible and invisible segments, and decouple objects of interest into static and dynamic subsets to isolate specific model behaviors. Tab.~\ref{tab:result-table} and Fig.~\ref{fig:example-results} summarize our results.
\newedit{The supplementary material provides model-inference details (Sec.~\ref{app:models}) and a human study assessing agreement with our metrics (Sec.~\ref{sec:human-study}). We also report category-wise results (Sec.~\ref{sec:category-wise-results}), temporal stability analysis (Sec.~\ref{sec:temporal-stability}), and an evaluation of 4DGT on longer videos (Sec.~\ref{sec:longer-videos}).}
We describe our key findings below.

\subsection{Key Findings}

\noindent\textbf{Seeing is not remembering.} 
A common hope is that if a 4D foundation model can effectively reconstruct or generate dynamic scenes from alternative viewpoints, it must inherently understand the underlying dynamics well enough to extrapolate beyond the observed frames. 
However, as shown in Tab.~\ref{tab:result-table} and Fig.~\ref{fig:result-figure}, all models suffer from a substantial performance drop on invisible segments compared to visible ones. \newedit{This drop suggests that visual memory is a capability distinct from reconstruction or novel-view synthesis when the target object remains observed.} This also suggests that current models still rely primarily on interpolating directly from observed inputs, rather than constructing an internal memory of the world that allows them to persist through space and time.

\noindent\textbf{Explicit conditioning helps visual memory.} 
Another observation is that models which explicitly project input observations onto target views before inpainting (GEN3C~\cite{gen3c}, TrajectoryCrafter~\cite{trajcrafter}, and NeoVerse~\cite{neoverse}) generally exhibit stronger visual memory. Notably, they significantly outperform others on invisible segments. We conjecture this is because most 4D models are predominantly trained on videos where target objects remain continuously visible~\cite{duan2026liveworldsimulatingoutofsightdynamics,ma2026sightmindevaluatingstate}. Consequently, when tasked with generating invisible segments, a scenario rarely encountered during training, they fall short. In contrast, explicit memory conditioning provides necessary guidance, reducing the complex generation process to a simpler refinement task. 

\noindent\textbf{Memory is multi-faceted.}
While the overall trends hold (\ie, models ``remember'' better on visible segments and static objects), each model excels at different aspects of visual memory. Using our complementary metrics, we are able to dissect their distinct strengths and weaknesses. For instance, as shown in Fig.~\ref{fig:result-figure}, while GEN3C~\cite{gen3c} and 4DGT~\cite{xu20254dgt} both excel at modeling motion continuity, GEN3C~\cite{gen3c} is significantly better at maintaining object permanence. Similarly, while NeoVerse~\cite{neoverse} and Imagine360~\cite{imagine360} track dynamics equally well, NeoVerse~\cite{neoverse} is much better at preserving object identity.

\noindent\textbf{Static memory is predictive of dynamic memory.}
While models generally remember static objects better than dynamic ones (Tab.~\ref{tab:result-table}), the results for both subsets are strongly correlated across all evaluated metrics (Fig.~\ref{fig:correlation}(d)). This suggests that models may rely on shared underlying mechanisms for both object types. Consequently, improvements in static memory are likely to translate to dynamic memory, shedding light on future directions for data curation and model refinement.

\noindent\textbf{Properly curated 360$^\circ$ data enhances visual memory.}
Since 360$^\circ$ videos capture the entire environment, it is natural to assume that models trained to predict the full panorama will inherently learn to remember. 
\newedit{However, this is only partly true. As shown in Tab. \ref{tab:result-table}, 360$^\circ$ models, specifically Argus and CubeComposer, perform well on the static subset, while suffering from large performance drops on the dynamic subset.}
We conjecture this is because they are predominantly trained on scenes with few dynamic objects~\cite{viewpoint,imagine360} or simple camera trajectories~\cite{cubecomposer}, rendering complex dynamic evaluation scenarios out-of-distribution. 
Within the group, Argus \cite{argus}, in contrast, is trained on more dynamic 360$^\circ$ scenes and explicitly models camera noise during training, and thus achieves the highest \texttt{Perm} and \texttt{Cont} scores among all models on the static subset, despite being built on a relatively old pretrain model~\cite{blattmann2023stable}. This suggests that 360$^\circ$ data is a powerful resource, not just for evaluation, but for actively cultivating visual memory in foundation models.

\input{figures/result_figure}

\input{figures/correlations.tex}

%% file: tables/main_table_v2.tex
\begin{table}[t]
    \centering
    \small
    \setlength{\tabcolsep}{3pt}
    \newlength{\maintableparenwidth}
    \settowidth{\maintableparenwidth}{00.00\%}
    \resizebox{\textwidth}{!}{
    \begin{tabular}{@{}lll*{6}{r}@{}}
    \toprule
    & \multirow{2}{*}{\raisebox{-0.6ex}{Models}} & \multirow{2}{*}{\raisebox{-0.6ex}{Memory}} & \multicolumn{3}{c}{Static Objects} & \multicolumn{3}{c}{Dynamic Objects} \\
    \cmidrule(lr){4-6}\cmidrule(lr){7-9}
    & & & \multicolumn{1}{c}{\makecell{\texttt{Permanence}}} & \multicolumn{1}{c}{\makecell{\texttt{Continuity}}} & \multicolumn{1}{c}{\makecell{\texttt{Appearance}}} & \multicolumn{1}{c}{\makecell{\texttt{Permanence}}} & \multicolumn{1}{c}{\makecell{\texttt{Continuity}}} & \multicolumn{1}{c}{\makecell{\texttt{Appearance}}}
    \\
    \midrule
    \multirow{4}{*}{\rotatebox[origin=c]{90}{4D Recon.}} & CUT3R \citep{cut3r} & Implicit & \phantom{0}3.14\% (66.43\%) & 70.66\% (86.76\%) & 20.97\% (56.71\%) & \phantom{0}0.99\% (43.99\%) & 58.27\% (76.94\%) & 18.04\% (49.49\%) \\
    & 4DGT \citep{xu20254dgt} & Explicit & 16.03\% (97.00\%) & 70.34\% (93.03\%) & 37.36\% (83.69\%) & \phantom{0}3.89\% (96.21\%) & 61.12\% (91.25\%) & 23.96\% (79.98\%) \\
    & CogNVS \citep{cognvs} & Explicit & 41.98\% (\secondtext{99.29\%}) & 67.91\% (96.10\%) & 64.08\% (92.65\%) & 61.99\% (\textfirst{99.84\%}) & \secondtext{61.53\%} (\secondtext{96.15\%}) & 57.14\% (\secondtext{92.75\%}) \\
    & NeoVerse \citep{neoverse} & Explicit & 78.92\% (97.23\%) & 63.66\% (85.84\%) & \secondtext{80.42\%} (91.77\%) & 54.31\% (88.04\%) & 48.01\% (71.33\%) & 65.54\% (83.08\%) \\
    \midrule
    \multirow{4}{*}{\rotatebox[origin=c]{90}{Cam. Ctrl.}} & ReCamMaster \citep{recammaster} & Implicit & 70.06\% (95.48\%) & 51.73\% (64.31\%) & 73.48\% (89.36\%) & 45.03\% (91.52\%) & 35.52\% (54.65\%) & 50.07\% (82.08\%) \\
    & TrajectoryCrafter \citep{trajcrafter} & Explicit & 81.56\% (\textfirst{99.55\%}) & 70.14\% (\textfirst{96.96\%}) & 77.52\% (\textfirst{96.30\%}) & \secondtext{77.80\%} (\secondtext{99.74\%}) & 56.45\% (\textfirst{97.01\%}) & \secondtext{66.74\%} (\textfirst{94.24\%}) \\
    & GEN3C \citep{gen3c} & Explicit & \secondtext{83.97\%} (95.96\%) & \secondtext{76.79\%} (86.98\%) & \textfirst{82.61\%} (\secondtext{93.26\%}) & \textfirst{87.06\%} (96.13\%) & \textfirst{64.62\%} (88.54\%) & \textfirst{72.28\%} (90.94\%) \\
    & HyDRA \citep{hydra} & Implicit & 70.38\% (\makebox[\maintableparenwidth][c]{N/A}) & 38.90\% (\makebox[\maintableparenwidth][c]{N/A}) & 65.65\% (\makebox[\maintableparenwidth][c]{N/A}) & 55.49\% (\makebox[\maintableparenwidth][c]{N/A}) & 35.41\% (\makebox[\maintableparenwidth][c]{N/A}) & 44.82\% (\makebox[\maintableparenwidth][c]{N/A}) \\
    \midrule
    \multirow{4}{*}{\rotatebox[origin=c]{90}{$360^\circ$ Gen.}} & ViewPoint \citep{viewpoint} & Implicit & 23.84\% (77.29\%) & 46.14\% (64.81\%) & 47.82\% (74.28\%) & 16.63\% (73.71\%) & 37.70\% (59.66\%) & 34.61\% (73.75\%) \\
    & Imagine360 \citep{imagine360} & Implicit & 34.17\% (98.24\%) & 53.78\% (93.93\%) & 40.78\% (83.24\%) & 25.84\% (98.14\%) & 44.86\% (94.30\%) & 22.76\% (80.09\%) \\
    & Argus \citep{argus} & Implicit & \textfirst{84.06\%} (99.25\%) & \textfirst{80.60\%} (95.86\%) & 68.49\% (87.69\%) & 62.42\% (98.97\%) & 59.86\% (95.22\%) & 49.31\% (85.20\%) \\
    & CubeComposer \citep{cubecomposer} & Implicit & 73.56\% (98.94\%) & 75.51\% (\secondtext{96.33\%}) & 68.42\% (93.18\%) & 45.29\% (99.27\%) & 52.95\% (95.33\%) & 37.86\% (87.55\%) \\
    \bottomrule
    \end{tabular}
    }
    \vspace{2pt}
    \caption{\newedit{\textbf{\methodname{} results.}}
    Experiments are conducted over multiple 4D reconstruction models (4D Recon.), camera-controllable video generation models (Cam. Ctrl.), and video-to-360$^\circ$ generation ($360^\circ$ Gen.) models.
    We report performance scores across static and dynamic subsets on both the invisible and visible (in parenthesis) segments.
    \textfirst{Green} denotes the highest score and \secondtext{yellow} denotes the second highest.
    Results for visible segments of HyDRA~\cite{hydra} are not applicable due to it only predicting frames after the object leaves view.}
    \vspace{-15pt}
    \label{tab:result-table}
\end{table}

%% file: figures/result_figure.tex
\begin{figure}[t]
    \centering
    \includegraphics[width=\linewidth]{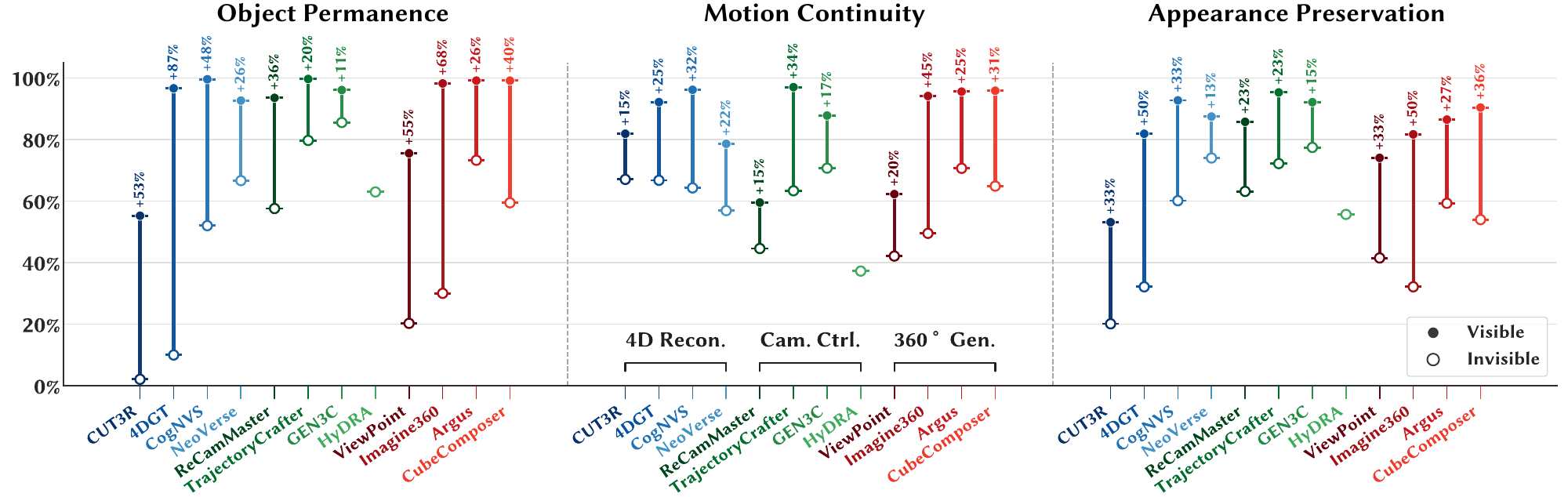}
    \vspace{-18pt}
    \caption{
    \textbf{\newedit{Performance gap between visible and invisible segments on both the dynamic and static subsets.}}
    We report, for each model and evaluation aspect, the drop in score from the visible to the invisible segment.
    Results for visible segments of HyDRA~\cite{hydra} are not applicable due to it only predicting frames after the object leaves view.
    All other models show a large gap across all aspects, suggesting that current models struggle to remember objects once they leave the input field of view.
    }
    \label{fig:result-figure}
  \end{figure}

%% file: figures/correlations.tex
\begin{figure}[t]
    \centering
    \includegraphics[width=\linewidth]{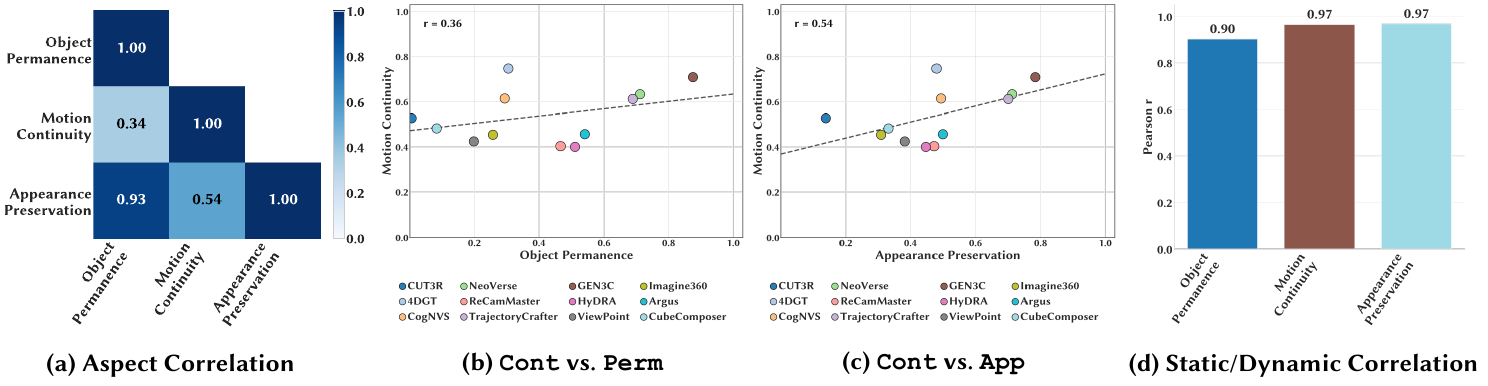}
    \vspace{-18pt}
    \caption{
    \textbf{\newedit{Correlation analysis.}}
    (a) The Pearson correlation matrix (computed on invisible segments across both static and dynamic subsets) shows that \texttt{Perm} and \texttt{App} are strongly correlated ($r=0.93$), while \texttt{Cont} is weakly correlated with each ($r=0.34$ and $r=0.54$).
    The \texttt{Cont}--\texttt{Perm} (b) and \texttt{Cont}--\texttt{App} (c) scatter plots confirm that motion continuity is a largely independent capability, suggesting that visual memory is multi-faceted and should not be reduced to a single metric.
    (d) Static and dynamic subset scores (on invisible segments) correlate highly across all aspects, indicating that memory for both static and dynamic objects draws on shared underlying capabilities.
    }
    \vspace{-8pt}
    \label{fig:correlation}
  \end{figure}

%% file: sections/7_limitations.tex
\section{Limitations}
\newedit{
Our evaluation relies on off-the-shelf models for both data construction and scoring. Camera poses estimated by ViPE~\cite{vipe} may contain errors, while pretrained models used for object matching, tracking, and scoring may introduce additional inaccuracies. These errors can affect the measured memory scores independently of the evaluated model's capabilities. 
Nonetheless, our ablations show that method rankings are largely preserved when replacing individual scoring components (Supp. Mat., Sec.~\ref{sec:off-the-shelf-ablations}), suggesting limited sensitivity to these model choices.
}

\newedit{
Additionally, our dataset is drawn from YouTube $360^\circ$ videos~\cite{DBLP:conf/nips/WallingfordBKRD24}. Although it covers diverse object categories, its scene distribution reflects the available source videos. Extending the data collection to other sources and less represented environments could broaden the benchmark's coverage.
}

%% file: sections/8_conclusion.tex
\vspace{-8pt}
\section{Conclusion}
\vspace{-5pt}
In this paper, we introduce \methodname{}, a benchmark for evaluating object-centric visual memory in 4D foundation models.
Our results reveal that while current models perform well at synthesizing novel views where most of the content is visible, they struggle to maintain object persistence once the object leaves the input field of view, indicating a fundamentally distinct capability.
We further identify a key contributor to this gap: current 4D foundation models are predominantly trained on videos in which target objects remain visible throughout, leaving little supervision signal for the model to learn regarding \textit{memory}.
Architectures that condition on explicit geometric structure partially compensate for this by encoding object persistence directly into the input, but it remains unclear whether such architectural advantages would persist under a memory-aware training paradigm.
Our analysis points to two complementary directions for future work: curating training data with frequent occlusion and reappearance, and further exploring architectures that combine explicit geometric conditioning with the generative priors of video models.
We believe \methodname{} can serve as a valuable testbed for driving progress toward 4D foundation models capable of true persistent memory, guiding future advances in both training and architecture design.

%% file: supp/supp.tex
\par\vskip 0.1in
\hrule height 4pt
\vskip 0.25in
\vskip -\parskip
{\centering
{\LARGE\bf Can 4D Foundation Models Remember?\\Supplementary Material\par}
\vskip 0.29in
\vskip -\parskip
\hrule height 1pt
\vskip 0.3in
}

\etocdepthtag.toc{appendix}
\begingroup
\etocsettagdepth{main}{none}
\etocsettagdepth{appendix}{subsection}
\etocsettocstyle{}{}
\tableofcontents{}
\endgroup
\clearpage

\input{supp/sections/0_dataset_curation}

\input{supp/sections/1_metrics}
\input{supp/sections/2_human_study}

\input{supp/sections/3_additional_experiments}
\input{supp/sections/4_models}
\input{supp/sections/5_visualization}

%% file: supp/sections/0_dataset_curation.tex
\section{Additional Details on Dataset Curation}
\label{app:dataset}

We provide additional details on each stage of the \methodname{} dataset construction pipeline: data collection and preprocessing, object discovery and tracking, and constrained camera trajectory generation.

\subsection{Data Collection and Preprocessing}

Following~\cite{argus,tu2026tldr}, we collect 360$^\circ$ videos from YouTube.
The downloaded videos undergo a filtering step to remove static clips, screen recordings, and low-quality footage before being segmented into 10-second clips.

\noindent\textbf{Camera pose estimation.}
For each clip, we crop its center region into a perspective video and run ViPE~\citep{vipe,tu2026tldr}, an optimization-based SLAM method, to obtain pseudo-ground-truth camera poses in metric scale.
Following~\citep{360anything}, we stabilize the estimated poses to maintain a consistent orientation within each clip.
To enforce gravity alignment, we uniformly sample 10 frames per clip, crop 8 perspective views per frame, and run GeoCalib~\citep{geocalib} to estimate the average gravity direction, which we then use to re-align all camera poses.

\subsection{Object Discovery and Tracking}

\noindent\textbf{Object identification.}
We run Qwen3-VL-Plus~\citep{qwen3vl} on 5 uniformly sampled frames from each 360$^\circ$ clip, prompting it to identify objects that match our construction requirements (static or dynamic).
The prompt explicitly distinguishes \emph{dynamic} from \emph{potentially movable} objects (\eg, a parked car is static), and we provide in-context examples to reduce ambiguity.
The model also outputs a category-level object description and, for dynamic objects, the motion type.
The full prompts used for dynamic and static object identification are shown in Fig.~\ref{fig:prompt-dynamic} and Fig.~\ref{fig:prompt-static}, respectively.

\begin{figure}[h]
\centering
\fbox{\begin{minipage}{0.93\linewidth}
\small
\textbf{Role:} You are an expert in identifying the dominant object in a video.

\smallskip
\textbf{Task:} Given a sequence of subsampled frames from a 10-second clip of a 360$^\circ$ video, identify the dominant \emph{dynamic} object in the video. If there are multiple dominant objects, identify a random dynamic object with visible motion.

\smallskip
\textbf{Instructions:}
\begin{enumerate}
    \item The images are subsampled frames from a 10-second 360$^\circ$ video clip.
    \item Look carefully at the images as well as how objects move throughout all the images.
    \item Identify the dominant object or the object with the most amount of visible motion.
    \item Separate object motion from camera motion. If the object remains static while the camera is moving, the object is \textbf{not} a dynamic object.
    \item Separate whether the object is \emph{movable} or \emph{dynamic}. A car is movable but not dynamic if it is parked, and should \textbf{not} be identified as a dynamic object.
    \item Classify the object into a short category description, e.g., ``car'', ``person'', ``animal'', ``bike''.
    \item Output your result strictly in the format below. If no dominant object is found, output \texttt{null} for all fields.
\end{enumerate}

\smallskip
\textbf{Output:} A JSON object with fields:
\texttt{"object\_name"} (string), \texttt{"object\_motion"} (specific motion such as ``driving'' or ``walking''; \texttt{"static"} if static; \texttt{null} if not found), \texttt{"object\_confidence"} (float).
\end{minipage}}
\caption{VLM prompt for dynamic object identification.}
\label{fig:prompt-dynamic}
\end{figure}

\begin{figure}[h]
\centering
\fbox{\begin{minipage}{0.93\linewidth}
\small
\textbf{Role:} You are an expert in identifying the static object in a video.

\smallskip
\textbf{Task:} Given a sequence of subsampled frames from a 10-second clip of a 360$^\circ$ video, identify a \emph{static} object in the video. If there are multiple static objects, identify a random one and describe its general category.

\smallskip
\textbf{Instructions:}
\begin{enumerate}
    \item The images are subsampled frames from a 10-second 360$^\circ$ video clip.
    \item If an object is certain not to move in real life, it is a static object.
    \item If an object is movable but not dynamic in this clip, it is still considered static (\eg, a parked car).
    \item The object should not be too large (\eg, ``ground'', ``sky'') nor too small to be visible.
    \item Classify the object into a short category description, e.g., ``building'', ``pillar''.
    \item Only name objects with clear, unambiguous edges. Do not describe objects like ``mountain'', ``snow'', or ``water''.
    \item Output your result strictly in the format below. If no dominant object is found, output \texttt{null} for all fields.
\end{enumerate}

\smallskip
\textbf{Output:} A JSON object with fields:
\texttt{"object\_name"} (string), \texttt{"object\_motion"} (\texttt{"static"}, or \texttt{null} if not found), \texttt{"object\_confidence"} (float).
\end{minipage}}
\caption{VLM prompt for static object identification.}
\label{fig:prompt-static}
\end{figure}

\noindent\textbf{360$^\circ$ tracking with SAM3.}
Given the object description, we prompt SAM3~\citep{sam3} to segment and track all matching instances throughout the 360$^\circ$ clip.
Directly applying SAM3 to equirectangular video fails when objects cross the left or right boundary: the object re-enters from the opposite edge and is often assigned a new identity, producing discontinuous tracks.

To address this, we horizontally concatenate a second copy of the same 360$^\circ$ clip to the right, so that edge-crossing trajectories remain continuous within the duplicated canvas.
This introduces duplicate tracks, which we remove using two rules:
\begin{enumerate}
    \item If a mask never touches the center seam or either boundary, a duplicate always exists in the copied half; we remove the instance in the right half.
    \item If a mask touches the center seam or either boundary, a duplicate exists that touches one of the boundaries; we keep the seam duplicate.
\end{enumerate}
Valid masks are then merged back to the original 360$^\circ$ frame coordinates.
We discard any track that begins after the first frame or is dropped mid-clip, yielding stable persistent segmentations for downstream trajectory generation.

\subsection{Constrained Camera Trajectory Generation}

\noindent\textbf{Object selection.}
For each clip, we rank tracked objects by their average mask area over time and select the top 10.
For each selected object we generate one paired perspective sequence (input and reference) via camera trajectory optimization.

\noindent\textbf{Visibility interval precomputation.}
A perspective camera with fixed FOV can be parameterized by yaw-pitch angles $(\theta_t, \phi_t)$ at frame $t$.
We precompute per-frame angle intervals for both axes:
$\Theta_t^{\mathrm{full}}$/$\Phi_t^{\mathrm{full}}$ are the ranges that make the object \emph{fully} visible, and $\Theta_t^{\mathrm{part}}$/$\Phi_t^{\mathrm{part}}$ are the ranges ensuring \emph{at least partial} visibility.
By definition, $\Theta_t^{\mathrm{full}} \subseteq \Theta_t^{\mathrm{part}}$ and $\Phi_t^{\mathrm{full}} \subseteq \Phi_t^{\mathrm{part}}$.
These intervals serve as frame-wise constraints during trajectory optimization.

\noindent\textbf{Visibility schedule.}
The input camera follows an \emph{invisible–visible–invisible} pattern: the object is unseen at the start and end, and briefly enters the frame during a middle window.
This middle visible window is what provides the model its only direct observation of the object; evaluating the invisible segments tests whether memory persists beyond that window.
We sample four key timesteps: (1) end of the first invisible segment, (2) start of the visible segment, (3) end of the visible segment, and (4) start of the final invisible segment.
The gaps between (1)–(2) and (3)–(4) allow smooth transitions between visibility states.
Concretely, we sample the visible segment length $T_\text{visible} \in [0.1T, 0.6T]$, its start time $t_\text{visible-start} \in [0.2T,\; 0.8T - T_\text{visible}]$, and the first invisible segment length $T_\text{first-invisible} \in [0.1\,t_\text{visible-start},\; 0.6\,t_\text{visible-start}]$.
The reference camera keeps the object visible at all times.

\noindent\textbf{Keyframe sampling.}
At each key timestep $t$, we sample $(\theta_t, \phi_t)$ conditioned on the desired visibility state.
For visible keyframes, we draw from a clipped Gaussian $\mathcal{N}(\mu, \sigma)$ inside the corresponding interval, with $\mu$ at the interval midpoint and $\sigma$ equal to one-quarter of the interval width.
For invisible keyframes, we sample uniformly outside the visibility interval.
All keyframe samples are constrained so the per-frame angular change does not exceed $1^\circ$ (i.e., $30^\circ$/s at 30 fps) and the total rotation within any window stays within $20^\circ$.
For invisible windows, we apply only $\theta$ (yaw) constraints in practice, as natural camera motion is predominantly horizontal.

\noindent\textbf{Trajectory optimization.}
Given sampled keyframes, we solve for dense trajectories $\{(\theta_t, \phi_t)\}_{t=1}^{T}$ via constrained optimization.
We initialize piecewise trajectories between consecutive keyframes using B-splines~\citep{bsplines} and optimize control points with L-BFGS-B~\citep{byrd1995limited}.
Let $x_t \in \{\theta_t, \phi_t\}$ denote the optimization variable at frame $t$.
The full objective is:
\begin{equation}
\begin{aligned}
\mathcal{L} =\;&
\lambda_{\mathrm{vis}}\!\left(\mathcal{L}_{\mathrm{inc}} + \mathcal{L}_{\mathrm{exc}}\right)
+ \lambda_{\mathrm{smooth}}\!\left(\mathcal{L}_{\mathrm{vel}} + \mathcal{L}_{\mathrm{acc}} + \mathcal{L}_{\mathrm{jerk}}\right) \\
&+ \lambda_{\mathrm{cont}}\!\left(\mathcal{L}_{\mathrm{pos\text{-}cont}} + \mathcal{L}_{\mathrm{vel\text{-}cont}} + \mathcal{L}_{\mathrm{acc\text{-}cont}}\right)
+ \lambda_{\mathrm{center}} \mathcal{L}_{\mathrm{center}},
\end{aligned}
\end{equation}
where:
\begin{itemize}
    \item $\mathcal{L}_{\mathrm{inc}}$ penalizes (via L1 distance to interval limit) any visible frame where $(\theta_t, \phi_t)$ falls outside $\Theta_t^{\mathrm{full}} \times \Phi_t^{\mathrm{full}}$.
    \item $\mathcal{L}_{\mathrm{exc}}$ penalizes any invisible frame where $(\theta_t, \phi_t)$ falls inside $\Theta_t^{\mathrm{part}} \times \Phi_t^{\mathrm{part}}$.
    \item $\mathcal{L}_{\mathrm{vel}}$, $\mathcal{L}_{\mathrm{acc}}$, $\mathcal{L}_{\mathrm{jerk}}$ are finite-difference smoothness penalties on velocity, acceleration, and jerk of $\{x_t\}$.
    \item $\mathcal{L}_{\mathrm{pos\text{-}cont}}$, $\mathcal{L}_{\mathrm{vel\text{-}cont}}$, $\mathcal{L}_{\mathrm{acc\text{-}cont}}$ enforce position, velocity, and acceleration continuity at spline junctions.
    \item $\mathcal{L}_{\mathrm{center}}$ is a weak centering regularizer that pulls visible frames toward the midpoint of the visibility interval, improving optimization stability.
\end{itemize}
Trajectories that violate any visibility constraint after optimization are rejected.

\subsection{Post-Processing and Quality Filtering}

\newedit{
\newedit{\noindent\textbf{Trajectory post-processing.}}
We further smooth the optimized camera trajectories before rendering the final video pairs. Concretely, we apply a low-pass filter to the per-frame camera angle sequences in the frequency domain, removing high-frequency oscillations while preserving the overall motion envelope. This is done for both the reference and input views after camera trajectory generation, which reduces residual jitter without changing the downstream mask-based validity checks.
}

\noindent\textbf{Mask refinement.}
Because masks projected from the equirectangular 360$^\circ$ segmentation can be low-resolution, we refine object masks on the generated video pairs using SAM2~\citep{sam2}.
Specifically, we run SAM2 on the reference video (where the object is always visible) using the projected 360$^\circ$ mask on the first frame as a prompt, and then project the refined reference masks onto the input video.
We discard any sequence in which the refined SAM2 mask and the original projected SAM3 mask do not overlap on any frame.

\noindent\textbf{Scale filtering.}
We also filter sequences with extreme object scale based on the minimum and maximum mask area over time: a sequence is dropped if the largest-frame mask exceeds 70\% of image pixels or if the smallest-frame mask falls below 2\% of image pixels.

\newedit{
\subsection{Segment Duration Statistics}
\label{app:segment-duration-statistics}
Fig.~\ref{fig:visibility-segment-durations} summarizes the durations of visible and invisible segments for the static and dynamic subsets. 
The median visible duration is $2.62$\,s in both subsets, while the median invisible duration is $2.50$\,s for dynamic objects and $2.62$\,s for static objects.
}
\input{supp_figures/visibility_segment_durations}

%% file: supp_figures/visibility_segment_durations.tex
\begin{figure}[htbp]
    \centering
    \includegraphics[width=0.9\linewidth]{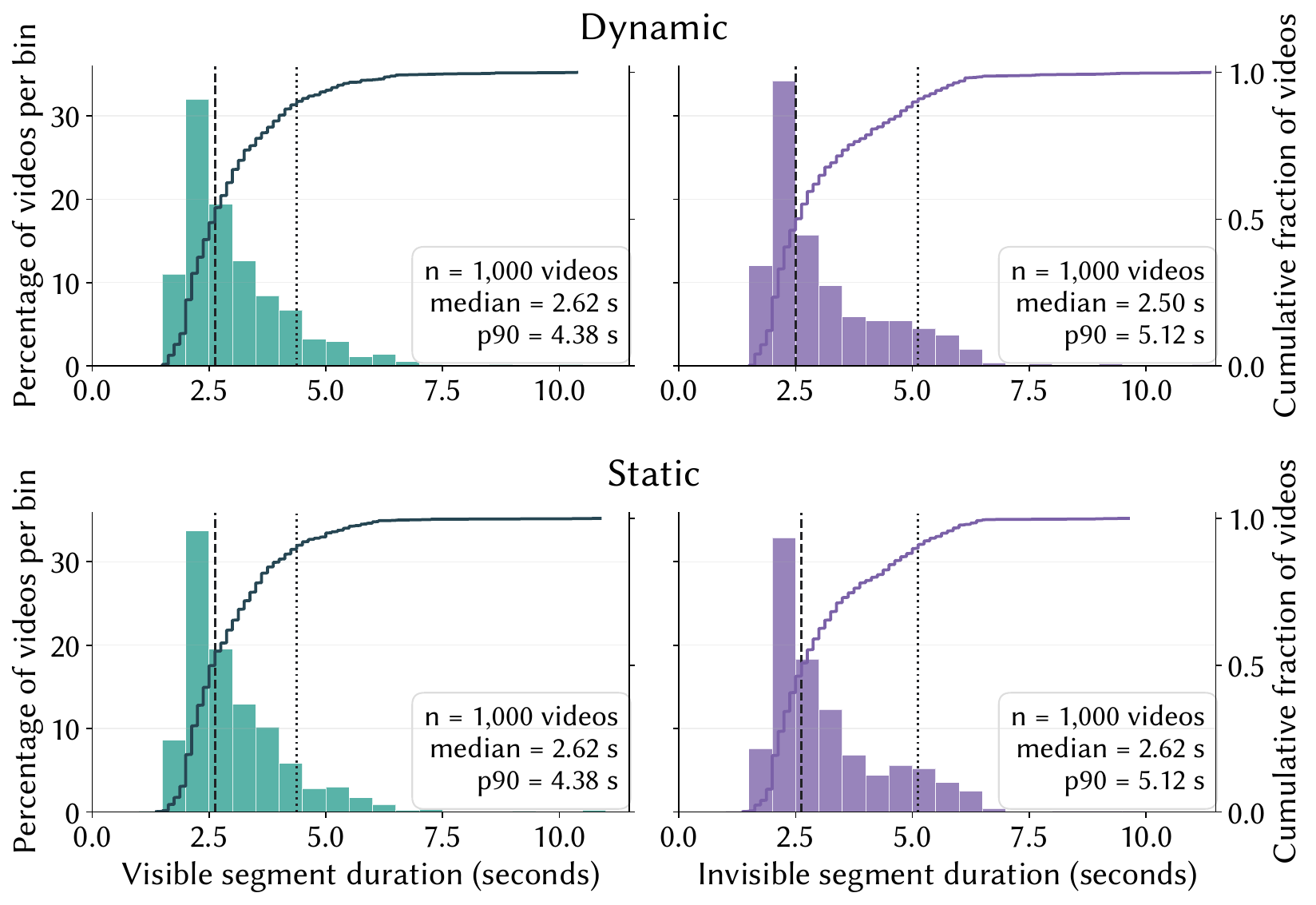}
    \caption{\newedit{\textbf{Visible and invisible segment durations for both subsets.} Distributions are shown for 1,000 dynamic object videos (top) and 1,000 static object videos (bottom), with visible segments on the left and invisible segments on the right. Bars show the percentage of videos per bin, and curves show the cumulative fraction of videos. Dashed and dotted vertical lines mark the median and 90th percentile, respectively. Durations are measured in seconds.}}
    \label{fig:visibility-segment-durations}
    \vspace{-15pt}
\end{figure}

%% file: supp/sections/1_metrics.tex
\section{Metric Implementation Details}
All metrics are computed after resizing the target-view videos and masks to a common evaluation resolution of $384 \times 512$ pixels. We first identify the first frame in which the target object is visible according to the trusted visibility annotation. Frames before this point are excluded from the final averages, since they do not provide a reliable target-object observation from which to initialize tracking.

Because model outputs do not include object masks, we infer predicted masks before applying the metrics defined in Sec.~\ref{sec:obj}. At the first visible frame, we crop the reference object using its reference mask and search for the most similar region in the corresponding generated frame using dense DINO feature matching. The reference mask support is then geometrically transferred into the matched predicted bounding box to form an initial predicted mask. Starting from this initialization, SAM2 propagates the mask through the remaining generated frames. A frame is treated as trackable if SAM2 returns a non-empty mask; otherwise, the predicted mask is empty and the SAM2 indicator for that frame is set to $0$.

\newedit{For the VLM component of \texttt{Perm}, the identity reference for each queried frame is produced by cropping the same-time-step reference frame using the reference mask bounding box with 10\% padding and resizing the result to $192 \times 256$. The generated frame is cropped using its SAM2-predicted mask bounding box with 10\% padding, resized to the same size, and tagged with a frame identifier. The prompt asks the VLM to judge whether the same semantic target object appears, without requiring matching appearance and while ignoring spatial alignment. Changes in pose, scale, lighting, texture, color, shape details, partial viewpoint, and moderate cropping are allowed as long as the object remains recognizable. The full prompt is shown in Fig.~\ref{fig:prompt-vlm-permanence}.}

\begin{figure}[h]
\centering
\fbox{\begin{minipage}{0.93\linewidth}
\small
\newedit{
\textbf{Role:} You are judging whether the target object in a prediction crop is the same object category/instance as the target object in a reference crop.

\smallskip
\textbf{Input description:} The first image is the reference object crop tagged as \texttt{REF}. It comes from the GT frame at the same time step as the prediction frame and is cropped using the target object's GT mask bounding box with 10\% padding. The second image is a prediction crop tagged with its unique frame ID, produced using the SAM2-predicted mask bounding box with 10\% padding.

\smallskip
\textbf{Internal reasoning:} Use chain-of-thought internally to identify and describe the target object in \texttt{REF}, compare the tagged prediction crop against the reference, and decide whether the object is visible in that crop without requiring matching appearance.

\smallskip
\textbf{Judging rules:}
\begin{enumerate}
    \item It is OK if the object changes pose, scale, lighting, texture, color, shape details, generated appearance, partial viewpoint, or is moderately cropped.
    \item Count it as visible as long as it can reasonably be recognized as the same semantic target object.
    \item If the prediction crop is empty, shows only background, shows a different object, or is too ambiguous, mark it not visible.
    \item Use appearance only as a weak clue, and ignore whether the crop is spatially aligned with the reference frame.
    \item Evaluate only the target object from \texttt{REF}.
\end{enumerate}

\smallskip
\textbf{Output:} Return exactly:
\begin{quote}
\ttfamily\small
<thinking>\\
short reasoning\\
</thinking>\\
<answer>\\
\{\\
\quad "reference\_description": "<brief description of the object in REF>",\\
\quad "tag": "\{frame\_tag\}",\\
\quad "visible": true,\\
\quad "reason": "<short reason>"\\
\}\\
</answer>
\end{quote}

\smallskip
\textbf{Output rules:} The JSON inside \texttt{<answer>} must be valid JSON. The frame tag must exactly match the queried frame tag, and \texttt{visible} must be a JSON boolean.
}
\end{minipage}}
\caption{VLM prompt for object permanence judging.}
\label{fig:prompt-vlm-permanence}
\end{figure}

\begin{figure}[h]
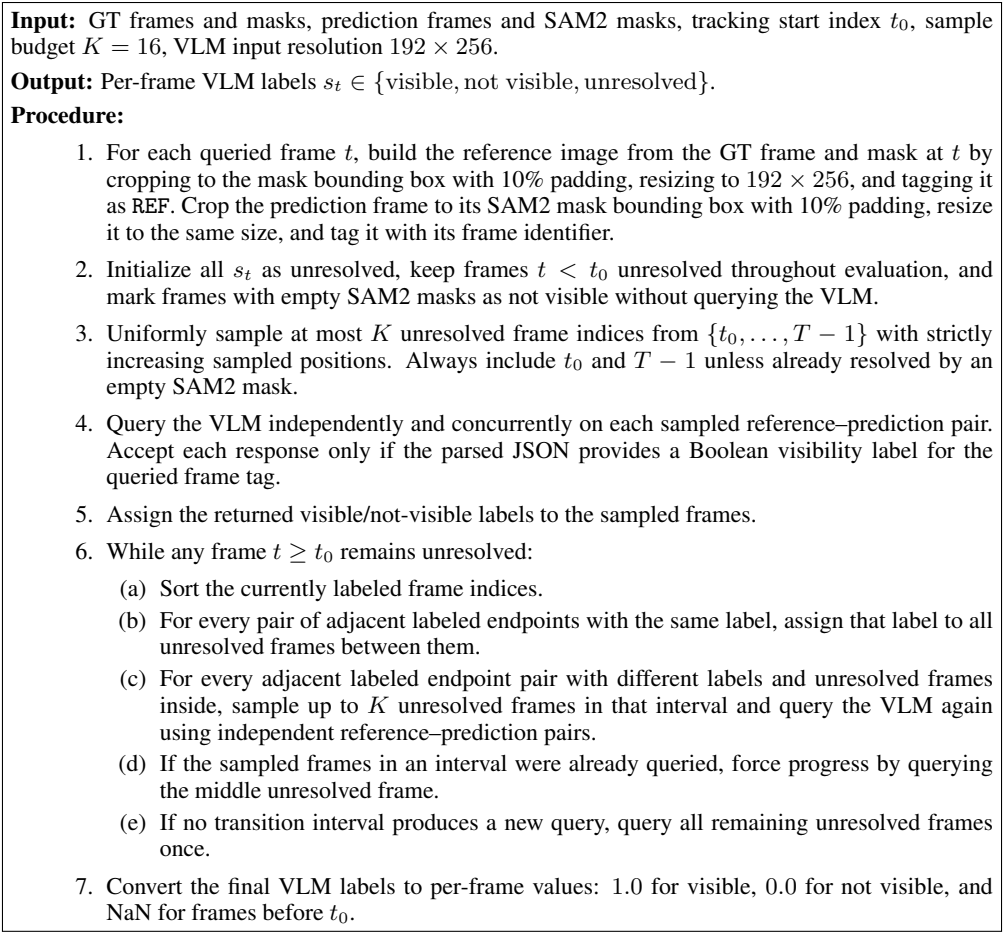

\centering
\fbox{\begin{minipage}{0.93\linewidth}
\small
\newedit{
\textbf{Input:} GT frames and masks, prediction frames and SAM2 masks, tracking start index $t_0$, sample budget $K=16$, VLM input resolution $192 \times 256$.

\smallskip
\textbf{Output:} Per-frame VLM labels $s_t \in \{\mathrm{visible}, \mathrm{not\ visible}, \mathrm{unresolved}\}$.

\smallskip
\textbf{Procedure:}
\begin{enumerate}
    \item For each queried frame $t$, build the reference image from the GT frame and mask at $t$ by cropping to the mask bounding box with 10\% padding, resizing to $192 \times 256$, and tagging it as \texttt{REF}. Crop the prediction frame to its SAM2 mask bounding box with 10\% padding, resize it to the same size, and tag it with its frame identifier.
    \item Initialize all $s_t$ as unresolved, keep frames $t < t_0$ unresolved throughout evaluation, and mark frames with empty SAM2 masks as not visible without querying the VLM.
    \item Uniformly sample at most $K$ unresolved frame indices from $\{t_0,\ldots,T-1\}$ with strictly increasing sampled positions. Always include $t_0$ and $T-1$ unless already resolved by an empty SAM2 mask.
    \item Query the VLM independently and concurrently on each sampled reference--prediction pair. Accept each response only if the parsed JSON provides a Boolean visibility label for the queried frame tag.
    \item Assign the returned visible/not-visible labels to the sampled frames.
    \item While any frame $t \geq t_0$ remains unresolved:
    \begin{enumerate}
        \item Sort the currently labeled frame indices.
        \item For every pair of adjacent labeled endpoints with the same label, assign that label to all unresolved frames between them.
        \item For every adjacent labeled endpoint pair with different labels and unresolved frames inside, sample up to $K$ unresolved frames in that interval and query the VLM again using independent reference--prediction pairs.
        \item If the sampled frames in an interval were already queried, force progress by querying the middle unresolved frame.
        \item If no transition interval produces a new query, query all remaining unresolved frames once.
    \end{enumerate}
    \item Convert the final VLM labels to per-frame values: $1.0$ for visible, $0.0$ for not visible, and NaN for frames before $t_0$.
\end{enumerate}
}
\end{minipage}}
\caption{Adaptive VLM labeling algorithm for the object permanence judge.}
\label{alg:vlm-permanence-judge}
\end{figure}

\newedit{
\noindent\textbf{Recursive VLM labeling.}
The VLM judge is applied adaptively, with independently queried frame pairs, rather than querying every generated frame. The full procedure is shown in Alg.~\ref{alg:vlm-permanence-judge}. Each pairwise VLM call is retried up to three times and accepted only when the parsed JSON provides a Boolean visibility label for the queried frame tag.
}

For \texttt{Cont}, invalid masks are handled before aggregation: frames with an empty reference mask, an empty predicted mask, or a failed SAM2 track are marked as invalid and skipped. The reference area used for scale normalization is the average mask area over the evaluated window after the tracking start, excluding invalid reference masks. The active implementation uses only image-plane center distances; camera-aware reprojection and velocity-distribution diagnostics exist in the evaluation code but are not used for the benchmark scores.

For \texttt{App}, DINO features are extracted from the full reference and generated frames after resizing each frame so that its height and width are multiples of the DINO patch size. The reference and predicted masks are downsampled to the DINO patch grid using area interpolation, and positive mask locations select the object tokens used for masked average pooling. If a mask selects no patch tokens, all patch tokens are used as a fallback for that frame. Auxiliary CLIP, DreamSim, DINO CLS-token, crop-based DINO, and max-pooling diagnostics are computed for analysis, but they are not included in the reported benchmark table.

\section{Metric Design Ablations}

\subsection{Appearance Score Design}
\label{sec:appearance-score-ablation}
We conduct an evaluation study to compare candidate appearance metrics and validate the final global DINO average-pooling design used for \texttt{App}. We consider crop-based CLIP similarity, crop-based DreamSim similarity~\citep{fu2023dreamsim}, DINO CLS-token similarity from object crops, DINO average patch-token similarity from crop features, and DINO max-pooled patch-token similarity from crop features. We also compare two global variants: global DINO average pooling, which first runs DINO on the full image and then average-pools patch tokens inside the object mask, and global DINO max pooling, which uses the same full-image features but aggregates the masked patch tokens with max pooling.

To quantify which metric best separates high-quality and low-quality object appearances, we sample 2,000 prediction-reference pairs from a model output and compute their object IoU, equivalently the $\mathcal{J}$-measure. We then extract the top 100 and bottom 100 pairs according to IoU. Since IoU close to $1$ indicates strong contour agreement with the reference object, we treat the top-IoU set as a pseudo high-quality appearance set and the bottom-IoU set as a pseudo low-quality set. For each candidate appearance metric, we compute the Earth Mover's Distance (EMD) between its score distributions on these two sets. A larger EMD indicates that the metric better separates clearly good object renderings from clearly poor ones.

As shown in Fig.~\ref{fig:emd-ablation}, global DINO average pooling consistently gives the largest distribution gap across evaluated models, outperforming crop-based CLIP, DreamSim, crop-level DINO variants, and global DINO max pooling. This supports our final design choice: aggregating DINO patch features from the full image within the object mask provides a more robust appearance signal than relying only on cropped inputs or max-pooled patch responses.

\input{supp_figures/emd_ablation.tex}

\subsection{Object Permanence Score Design}
\label{sec:permanence-score-ablation}
We also ablate the dual-verification design of \texttt{Perm}. As described in Sec.~\ref{sec:obj}, object permanence should require more than a plausible image region: the target object must remain spatially coherent enough to track and semantically recognizable as the same object. We therefore combine a SAM2 tracking signal with a VLM identity signal.

Fig.~\ref{fig:perm-ablation} illustrates why both components are needed. In the first case, the target is a tree and SAM2 correctly indicates that the original tracked object has been lost. The VLM alone, however, can confuse other nearby trees with the reference tree and incorrectly mark the object as present. This shows why tracking is necessary: it detects discontinuities in the object instance that a semantic judge may overlook. In the second and third cases, SAM2 continues to track a heavily degraded object-like blob. The tracked region remains spatially coherent, but the object identity is no longer recognizable. The VLM correctly rejects these cases, showing that semantic identity checking is necessary when tracking alone follows a corrupted region.

These examples motivate the logical-AND design in \texttt{Perm}: SAM2 verifies instance-level continuity, while the VLM verifies object identity and recognizability. Their combination better matches the goal of measuring whether the same observed object persists in the generated sequence.

\input{supp_figures/perm_ablation.tex}

\newedit{
\subsection{Motion Continuity Score Design}
\label{sec:motion-continuity-ablation}
To examine whether our results depend on the centroid-based definition of \texttt{Cont}, we recompute motion continuity using three alternatives: bounding-box center L2 distance, mask IoU, and dynamic time warping (DTW) over the centroid trajectory. We reuse the existing tracked masks across all 12 evaluated methods on the same subset of 200 videos (100 dynamic and 100 static), changing only the motion continuity calculation. As shown in Tab.~\ref{tab:ablation-ranking-correlations}, Spearman's rank correlations with default \texttt{Cont} range from $0.979$ to $1.000$ on visible segments and from $0.944$ to $0.993$ on invisible segments. The method rankings are therefore largely preserved under all three alternatives, suggesting that our comparisons do not depend strongly on the specific choice of centroid distance.
}

\newedit{
\subsection{Tracking-Conditioned Aggregation}
\label{sec:tracking-conditioned-aggregation}
We compare the Spearman correlations of \texttt{Cont} and \texttt{App} with \texttt{Perm} under all-frame and tracked-frame aggregation in Tab.~\ref{tab:tracking-conditioned-correlations}, using the 12 methods over the full dataset. All-frame aggregation assigns zero to untracked frames, approximately doubling the correlation between \texttt{Cont} and \texttt{Perm}. This makes motion continuity more dependent on object permanence, contrary to our intended separation of these aspects. Our default tracked-frame aggregation preserves this distinction more effectively, while \texttt{App} remains strongly correlated with \texttt{Perm} under both settings.
}

\begin{table}[htbp]
    \centering
    \small
    \begin{tabular}{@{}lrr@{}}
    \toprule
    Aggregation & \texttt{Cont}--\texttt{Perm} & \texttt{App}--\texttt{Perm} \\
    \midrule
    All-frame & 0.685 & 0.972 \\
    Tracked-frame (default) & 0.343 & 0.965 \\
    \bottomrule
    \end{tabular}
    \vspace{5pt}
    \caption{\newedit{\textbf{Tracking-conditioned aggregation ablation.} Spearman's rank correlations between all-frame and tracked-frame aggregation across 12 methods and all 2,000 cases. All-frame aggregation assigns untracked frames with zero scores, whereas tracked-frame aggregation averages only successfully tracked frames.}}
    \label{tab:tracking-conditioned-correlations}
\end{table}

\newedit{
\subsection{Off-the-Shelf Model Ablations}
\label{sec:off-the-shelf-ablations}
We conduct these ablations across all 12 evaluated methods on a subset of 200 videos, comprising 100 dynamic and 100 static cases.
To examine how the choice of off-the-shelf models affects our evaluation, we replace the initial matching model, tracking model, VLM judge, and appearance feature extractor separately. Fig.~\ref{fig:off-the-shelf-ablation-ranking} shows the resulting rankings across the 12 evaluated methods for both visible and invisible segments. 
For initial matching, we replace DINOv3-L/16~\citep{dinov3} with DINOv2-B/14~\citep{dinov2}, an earlier and smaller model from the same family, and CLIP-L/14~\citep{clip} and SigLIP2-L/16~\citep{siglip2}, which use image-text training rather than DINO's image-only self-supervised training.
For tracking, we replace SAM2.1 Large~\citep{sam2} with SAM2.1 Tiny to test a smaller model within the same generation, SAM3.1~\citep{sam3} to test a newer generation of the SAM family, and XMem~\citep{xmem}, an earlier video object segmentation model with a different memory-based architecture.
For the VLM judge, we replace Qwen3.8-27B~\citep{qwen38} with Gemini 3.5 Flash~\citep{gemini35flash} and Claude Sonnet 5~\citep{claudesonnet5}, testing whether object-presence judgments remain consistent across models from different providers.
For appearance scoring, we replace DINOv2-L/14~\citep{dinov2} with DINOv2-B/14 to test model size, DINOv3-L/16 to test model generation, and CLIP-L/14 and SigLIP2-L/16 to test different feature families. We evaluate initial matching and appearance scoring separately to distinguish their effects on the final scores.
We also report Spearman's rank correlations between the default and variant configurations in Tab.~\ref{tab:ablation-ranking-correlations}, showing that method rankings are largely preserved across component replacements.
}

\input{supp_figures/off_the_shelf_ablation_ranking}

\newedit{
\subsection{Camera Perturbation Analysis}
\label{sec:camera-perturbation}
We conduct this analysis across all 12 evaluated methods on a subset of 200 videos, comprising 100 dynamic and 100 static cases.
Since our evaluation uses rendered outputs, errors in camera control can affect the measured memory scores. To show the effect of camera control capability on the metrics, we apply random shift sequences to the generated videos and recompute the three metrics while keeping the reference videos fixed. We compare the default setting with low and high perturbation levels corresponding to 3\% and 9\% of the image width. As shown in Fig.~\ref{fig:camera-ablation-ranking}, \texttt{Perm} and \texttt{App} rankings are largely preserved, while \texttt{Cont} shows slightly more ranking changes across all methods, consistent with its use of object position.
This shows that \texttt{Perm} and \texttt{App} are largely robust to model's camera control capability, while \texttt{Cont} can be slightly affected.
}

\input{supp_figures/camera_ablation_ranking}

\newedit{
\subsection{Rotation-Dominant Subset Analysis}
\label{sec:rotation-dominant-subset}
Video-to-$360^\circ$ methods produce target views using camera rotation alone, whereas other methods also account for target-camera translation. To examine whether this difference affects cross-method comparisons, we re-aggregate existing predictions on a rotation-dominant subset of 200 videos and compare the resulting rankings with the default rankings across all 12 methods. As shown in Tab.~\ref{tab:ablation-ranking-correlations}, the invisible-segment Spearman correlations are 0.895, 0.916, and 0.930 for \texttt{Perm}, \texttt{Cont}, and \texttt{App}, respectively. The corresponding visible-segment correlations are 0.825, 0.944, and 0.965. These results show that method rankings remain broadly consistent on the rotation-dominant subset, although some changes remain. This supports the robustness of the overall comparison when translation effects are reduced, without fully isolating camera handling from other differences between methods.
}

\newedit{
\subsection{Correlation with Existing Benchmarks}
\label{sec:worldscore-correlation}
To examine whether our metrics reflect general video quality, we also evaluate the generated outputs using WorldScore~\citep{duan2025worldscore}, which measures video consistency, quality, and dynamics. These metrics evaluate the generated video without comparing the target object with its reference observation. We compute Spearman's rank correlations between the six WorldScore metrics and our default \texttt{Perm}, \texttt{Cont}, and \texttt{App} scores across the 12 evaluated methods on a subset of 200 videos (100 dynamic and 100 static). 
As shown in Tab.~\ref{tab:ablation-ranking-correlations}, more than half of the correlations with WorldScore are zero or negative, indicating substantial differences between its method rankings and ours.
These results suggest that our reference-based, object-centric evaluation provides information beyond general video quality and consistency, rather than reproducing the rankings from existing metrics.
}

\input{tables/ablation_ranking_correlations}

\subsection{Reference Score}
Tab.~\ref{tab:reference-score} reports the score obtained by evaluating the reference videos with the same metric pipeline used for all model outputs. Values outside parentheses are measured on invisible frames, and values inside parentheses are measured on visible frames.

\begin{table}[h]
    \centering
    \small
    \setlength{\tabcolsep}{4pt}
    \resizebox{\textwidth}{!}{
    \begin{tabular}{@{}l*{6}{r}@{}}
    \toprule
    & \multicolumn{3}{c}{Static Objects} & \multicolumn{3}{c}{Dynamic Objects} \\
    \cmidrule(lr){2-4}\cmidrule(lr){5-7}
    & \multicolumn{1}{c}{\texttt{Perm}} & \multicolumn{1}{c}{\texttt{Cont}} & \multicolumn{1}{c}{\texttt{App}} & \multicolumn{1}{c}{\texttt{Perm}} & \multicolumn{1}{c}{\texttt{Cont}} & \multicolumn{1}{c}{\texttt{App}} \\
    \midrule
    Reference & 99.99\% (99.99\%) & 98.68\% (98.88\%) & 99.79\% (99.82\%) & 99.62\% (99.96\%) & 97.84\% (98.76\%) & 99.69\% (99.86\%) \\
    \bottomrule
    \end{tabular}
    }
    \vspace{1pt}
    \caption{\textbf{Reference scores on the full benchmark.} The reference video is evaluated by the same pipeline as model predictions.}
    \label{tab:reference-score}
\end{table}

The reference scores are close to, but not exactly, $100\%$ because the reference video is treated as a baseline prediction and passed through the same mask-inference pipeline as every model output. In particular, SAM2 is run on the reference video to produce tracked masks, and repeated SAM2 tracking introduces slight mask noise. This mask noise affects centroid-based \texttt{Cont} and masked-feature \texttt{App}, while \texttt{Perm} remains perfect. Since all reference scores remain very close to $100\%$, this small deviation is acceptable and reflects measurement noise rather than a failure of the reference video.

%% file: supp_figures/emd_ablation.tex
\begin{figure}[h]
    \centering
    \includegraphics[width=\linewidth]{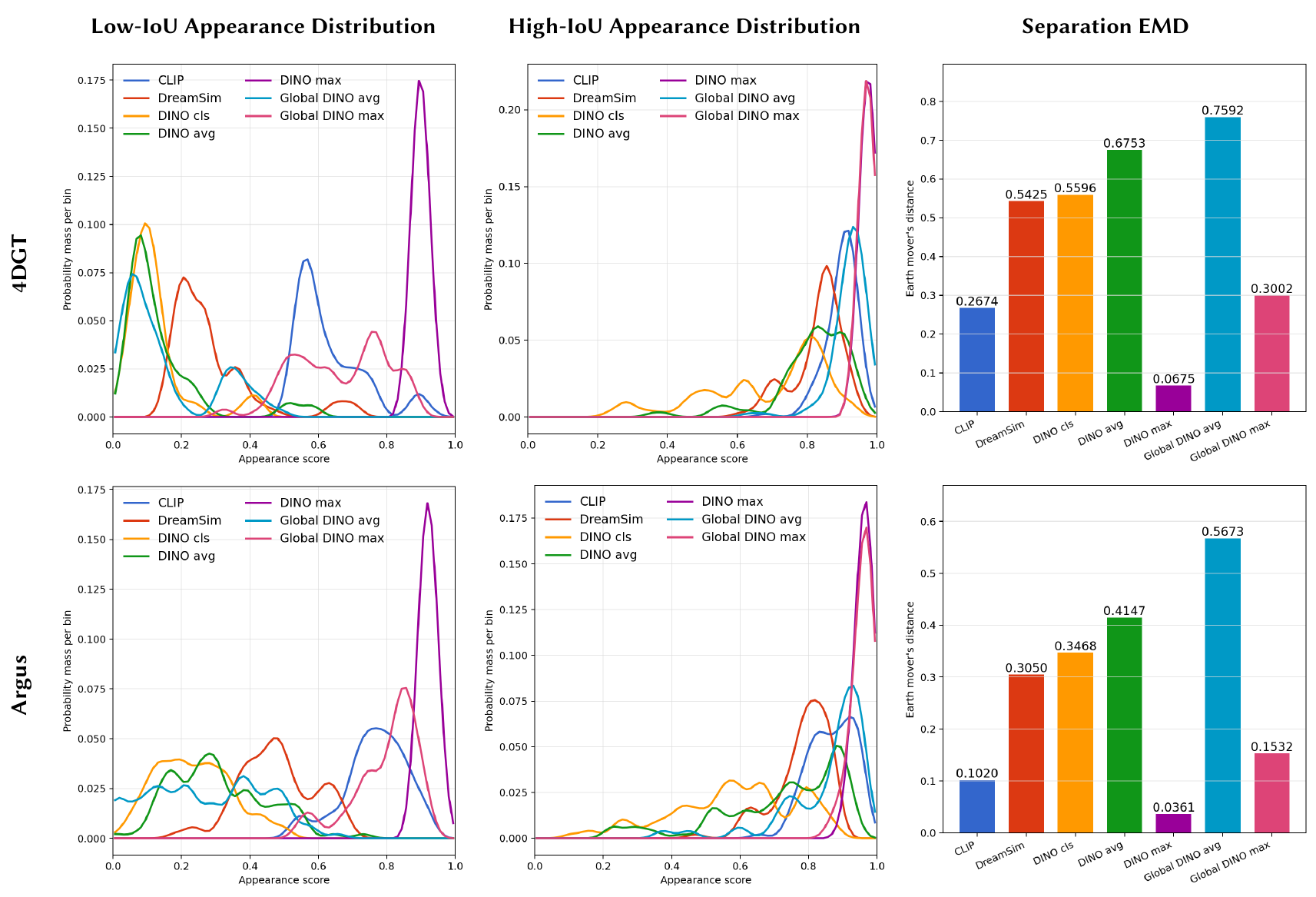}
    \caption{\textbf{Appearance metric ablation.} We rank prediction-reference pairs by object IoU and compare each metric's score distributions on the top and bottom 100 pairs using EMD. CLIP assigns high scores to both sets, DreamSim shows weak separation with degraded ordering, DINO CLS under-scores high-quality pairs, crop DINO average pooling separates reasonably but less strongly, and DINO max/global DINO max over-score both sets due to max aggregation. Global DINO average pooling gives the largest EMD and clearest separation, motivating its use in \texttt{App}.}
    \label{fig:emd-ablation}
\end{figure}

%% file: supp_figures/perm_ablation.tex
\begin{figure}[h]
    \centering
    \includegraphics[width=0.92\linewidth]{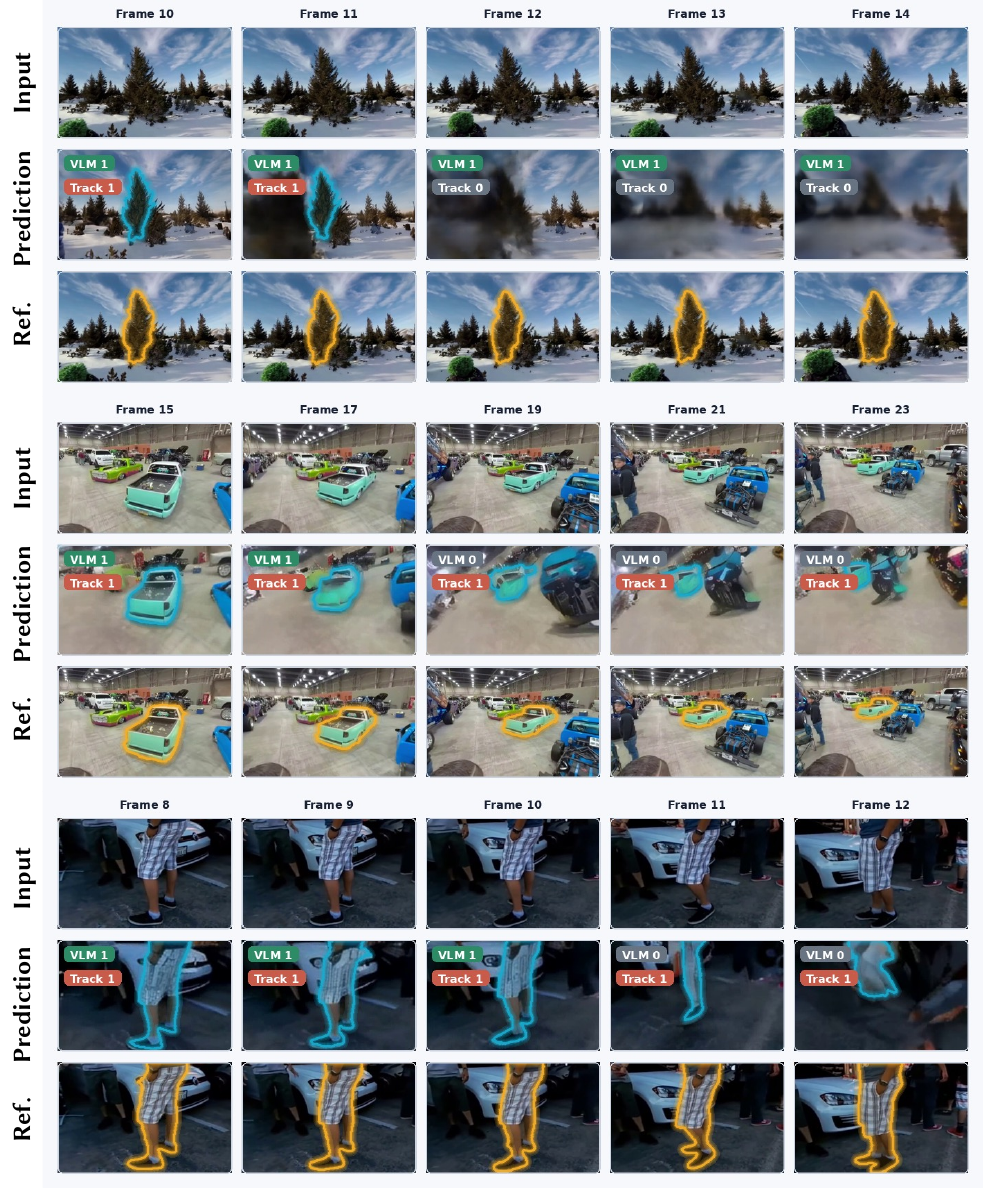}
    \caption{\textbf{Object permanence score ablation.} SAM2 tracking and VLM identity checking capture complementary failure modes. In case 1, SAM2 correctly detects that the tracked object is lost, while the VLM mistakes other trees for the target tree. In cases 2 and 3, SAM2 continues tracking a severely degraded object-like blob, but the VLM rejects it as no longer recognizable as the target object. Requiring both signals therefore better matches our object-centric definition of permanence.}
    \label{fig:perm-ablation}
\end{figure}

%% file: supp_figures/off_the_shelf_ablation_ranking.tex
\begin{figure}[t]
    \centering
    \includegraphics[width=\linewidth,height=0.82\textheight,keepaspectratio]{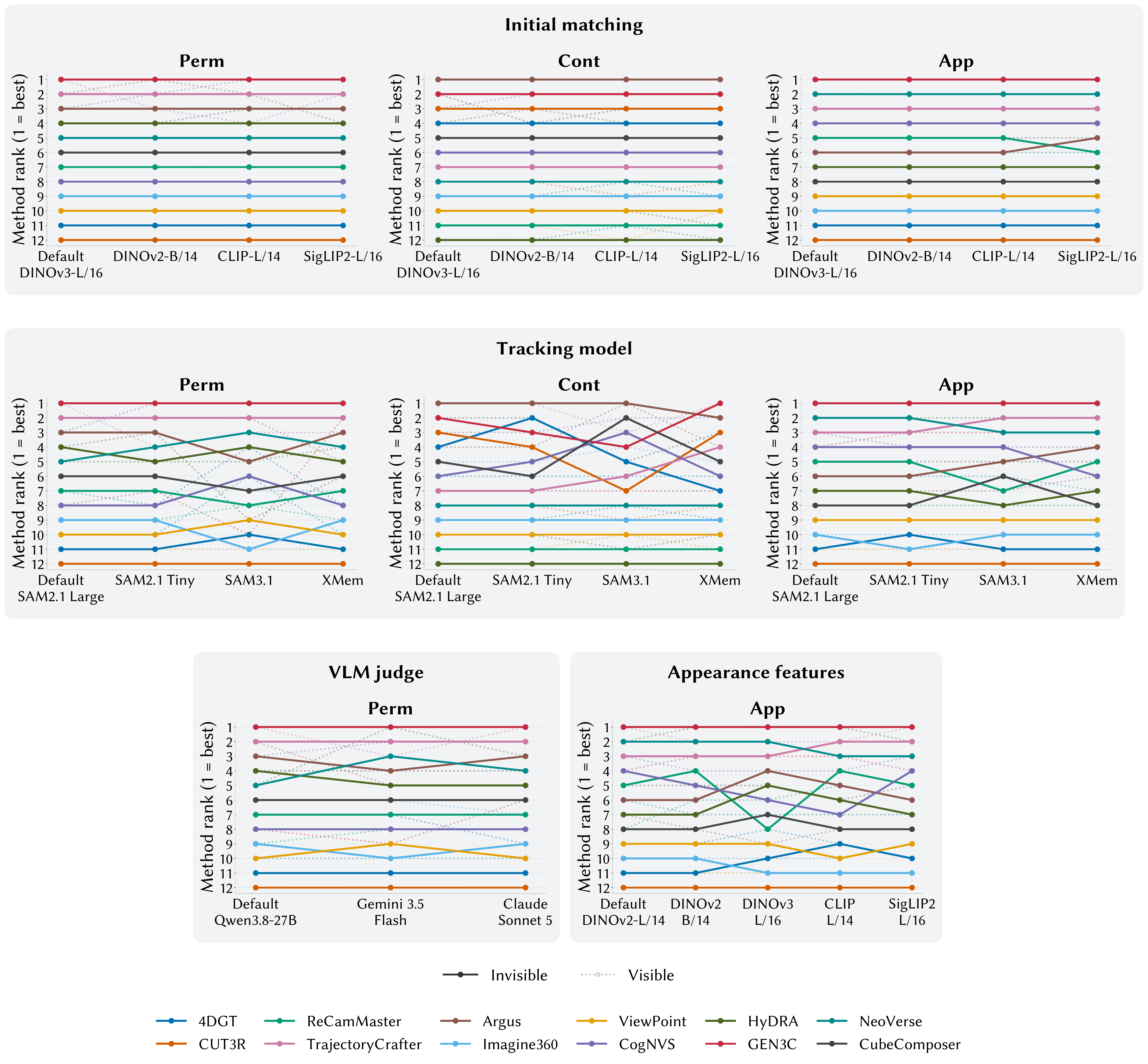}
    \caption{\newedit{\textbf{Effect of off-the-shelf model choices on method rankings.} We replace the initial matching model (top), tracking model (middle), VLM judge (bottom left), and appearance feature extractor (bottom right), and compare the rankings of the 12 evaluated methods with the default configuration on a subset of 200 videos (100 dynamic and 100 static). The default and replacement models are listed on each horizontal axis. Solid and dotted lines show invisible and visible segment rankings, respectively. Rank 1 denotes best.
    While changing the model slightly affects the method rankings, they are mainly preserved across largely different off-the-shelf models spanning diverse categories and performance tiers.}}
    \label{fig:off-the-shelf-ablation-ranking}
\end{figure}

%% file: supp_figures/camera_ablation_ranking.tex
\begin{figure}[t]
    \centering
    \includegraphics[width=\linewidth]{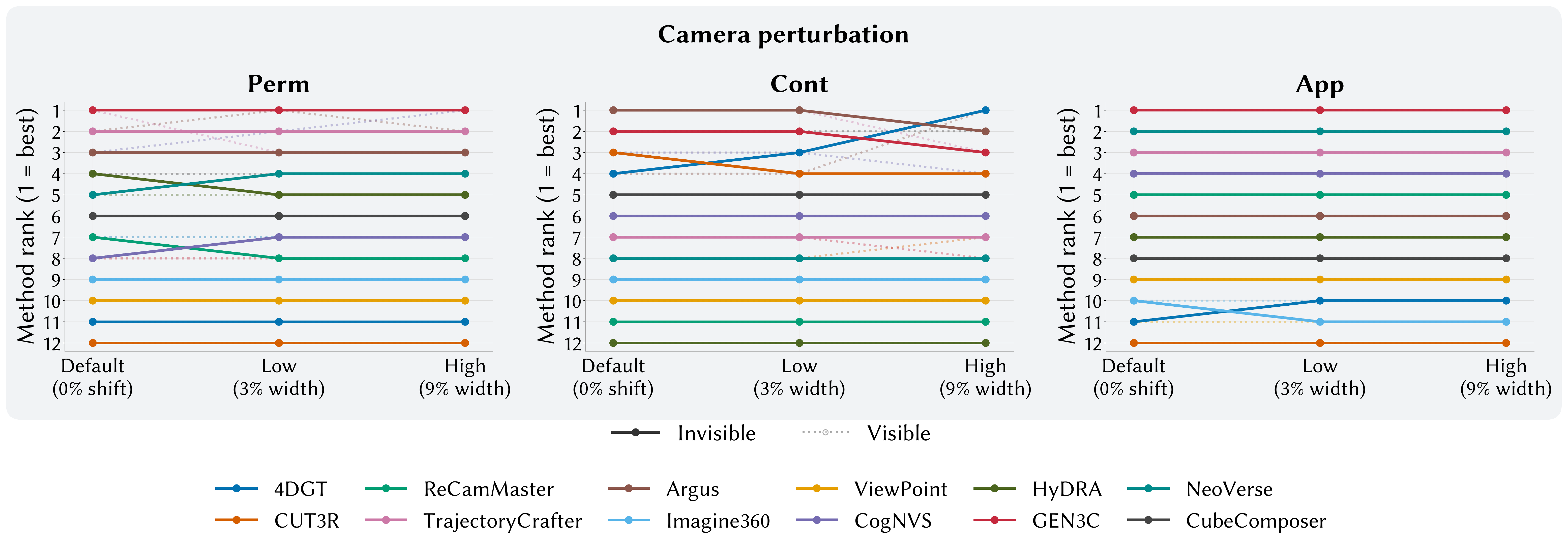}
    \caption{\newedit{\textbf{Effect of camera perturbation on method rankings.} We compare the rankings of the 12 evaluated methods on a subset of 200 videos (100 dynamic and 100 static) under the default setting and low and high frame perturbation levels corresponding to 3\% and 9\% of the image width. Solid and dotted lines show invisible and visible segment rankings, respectively. Rank 1 denotes the best. Rankings are largely stable for \texttt{Perm} and \texttt{App}, and \texttt{Cont}.}}
    \label{fig:camera-ablation-ranking}
    \vspace{5pt}
\end{figure}

%% file: tables/ablation_ranking_correlations.tex
\begin{table}[t]
    \centering
    \small
    \setlength{\tabcolsep}{5pt}
    \resizebox{0.9\linewidth}{!}{%
    \begin{tabular}{llccc}
    \toprule
    Group & Variant & \texttt{Perm} & \texttt{Cont} & \texttt{App} \\
    \midrule
    \rowcolor{blue!6}
        & DINOv2-B/14      & 1.000 (0.979) & 1.000 (0.979) & 1.000 (1.000) \\
    \rowcolor{blue!6}
        & CLIP-L/14        & 1.000 (0.951) & 1.000 (0.979) & 1.000 (1.000) \\
    \rowcolor{blue!6}
    \multirow{-3}{*}{Initial matching}
        & SigLIP2-L/16     & 1.000 (0.965) & 1.000 (0.986) & 0.993 (1.000) \\
    \midrule
    \rowcolor{blue!6}
        & SAM2.1 Tiny      & 0.993 (0.944) & 0.972 (1.000) & 0.993 (0.993) \\
    \rowcolor{blue!6}
        & SAM3.1           & 0.930 (0.776) & 0.860 (0.902) & 0.958 (0.993) \\
    \rowcolor{blue!6}
    \multirow{-3}{*}{Tracking model}
        & XMem             & 0.993 (0.790) & 0.930 (0.993) & 0.965 (0.986) \\
    \midrule
    \rowcolor{blue!6}
        & Gemini 3.5 Flash & 0.972 (0.888) & ---           & ---           \\
    \rowcolor{blue!6}
    \multirow{-2}{*}{VLM judge}
        & Claude Sonnet 5  & 0.993 (0.916) & ---           & ---           \\
    \midrule
    \rowcolor{blue!6}
        & DINOv2-B/14      & ---           & ---           & 0.993 (0.965) \\
    \rowcolor{blue!6}
        & DINOv3-L/16      & ---           & ---           & 0.916 (0.951) \\
    \rowcolor{blue!6}
        & CLIP-L/14        & ---           & ---           & 0.930 (0.944) \\
    \rowcolor{blue!6}
    \multirow{-4}{*}{Appearance features}
        & SigLIP2-L/16     & ---           & ---           & 0.986 (0.972) \\
    \midrule
    \rowcolor{green!7}
        & BBox center L2   & ---           & 0.993 (0.993) & ---           \\
    \rowcolor{green!7}
        & Mask IoU         & ---           & 0.944 (1.000) & ---           \\
    \rowcolor{green!7}
    \multirow{-3}{*}{Motion continuity definition}
        & DTW              & ---           & 0.986 (0.979) & ---           \\
    \midrule
    \rowcolor{orange!10}
        & Low (3\% width)  & 0.986 (0.979) & 0.993 (1.000) & 0.993 (1.000) \\
    \rowcolor{orange!10}
    \multirow{-2}{*}{Camera perturbation}
        & High (9\% width) & 0.986 (0.972) & 0.958 (0.944) & 0.993 (1.000) \\
    \midrule
    \rowcolor{violet!7}
    Evaluation subset
        & Rotation dominant & 0.895 (0.825) & 0.916 (0.944) & 0.930 (0.965) \\
    \midrule
    \rowcolor{gray!10}
        & 3D Consistency          & 0.490 (0.175)   & -0.231 (-0.161) & 0.524 (0.559)   \\
    \rowcolor{gray!10}
        & Photometric Consistency & -0.063 (-0.308) & 0.000 (-0.448)  & -0.049 (0.021)  \\
    \rowcolor{gray!10}
        & Style Consistency       & -0.105 (0.105)  & 0.524 (0.371)   & -0.294 (-0.280) \\
    \rowcolor{gray!10}
        & Subjective Quality      & 0.497 (0.179)   & -0.382 (-0.095) & 0.550 (0.634)   \\
    \rowcolor{gray!10}
        & Motion Magnitude        & 0.175 (0.503)   & -0.133 (0.427)  & -0.021 (0.147)  \\
    \rowcolor{gray!10}
    \multirow{-6}{*}{WorldScore~\citep{duan2025worldscore}}
        & Motion Smoothness       & -0.245 (-0.517) & -0.063 (-0.559) & -0.049 (-0.119) \\
    \bottomrule
    \end{tabular}%
    }
    \vspace{5pt}
    \caption{\newedit{\textbf{Ranking correlations under off-the-shelf model replacements, motion continuity definitions, camera perturbations, rotation-dominant subset selection, and with WorldScore~\citep{duan2025worldscore}.} We report Spearman's $\rho$ between the default and variant rankings across the 12 evaluated methods for each metric. All experiments use 200 sampled videos. 
    Off-the-shelf variants replace the initial matching model, tracking model, VLM judge, and appearance feature extractor separately. 
    The motion continuity variants reuse the existing masks and compare bounding-box center L2 distance, mask IoU, and dynamic time warping (DTW) over centroid trajectories with default \texttt{Cont}. 
    Camera perturbations apply low and high shifts of 3\% and 9\% of the image width. 
    The rotation-dominant row compares default rankings with those obtained on the rotation-dominant subset. 
    The WorldScore rows report correlation between each WorldScore metric and our default metric rankings. 
    Values outside parentheses are measured on invisible segments, and values inside parentheses are measured on visible segments. 
    Dashes indicate metrics not affected by the corresponding ablation.}}
    \label{tab:ablation-ranking-correlations}
\end{table}

%% file: supp/sections/2_human_study.tex
\newedit{
\section{Human Study}
\label{sec:human-study}
We conduct a human study to compare our automatic memory metrics with human judgments of object permanence, motion continuity, and appearance preservation across the 12 evaluated methods.

\subsection{Study Protocol}
In each trial, participants compare two videos generated by different methods for the same input, alongside the reference video and a copy with the target object highlighted in cyan. The generated videos are labeled A and B, with method names and automatic scores hidden. Controls allow participants to pause, replay, move through the videos, and adjust playback speed. Participants are asked to compare each memory property by selecting \emph{A better}, \emph{Tie}, \emph{B better}, or \emph{Not sure}.

The study covers all 66 pairs of methods, with 24 distinct cases per pair covering both the dynamic and static subsets, as well as visible and invisible segments. 
We then randomly shuffle and divide these into 66 questionnaires, each containing 24 main trials, two introductory examples, one gold-standard control trial, and one repeated main trial with the positions of A and B swapped. 

We exclude a submitted questionnaire if the participant selects the generated video over the reference for any metric in the gold-standard trial, or selects different videos for the same metric in the original and repeated trials. 
After collecting and filtering, 864 main comparisons remain.

\subsection{Participant Instructions}
Fig.~\ref{fig:human-study-interface} shows the participant instructions and an example case. Participants are asked to focus on the cyan-highlighted target object and ignore the background, other objects, people, camera quality, and unrelated differences between the videos. Each trial uses the following three prompts:
\begin{enumerate}
    \item \textbf{Object permanence:} Which video contains the target object for longer? Judge only whether the object continues to exist. Ignore how it looks or moves.
    \item \textbf{Motion continuity:} Does the object follow similar motion to the target object in the reference video? If the target object is static, its position shouldn't change (up to camera movement). Appearance doesn't matter.
    \item \textbf{Appearance preservation:} Which one of the generated objects looks more similar to the target object in the reference video? Motion doesn't matter.
\end{enumerate}
}

\input{supp_figures/human_study_interface}

\newedit{
\subsection{Agreement with Automatic Metrics}
For each metric, we compare human judgments with our default \methodname{} scores for the same cases. We include a comparison only when the human response is A, B, or a tie and both methods have valid automatic scores. 

We assign one point for a win, half a point for a tie, and zero for a loss, for both human judgments and automatic scores. 
For automatic comparisons, the method with the higher score wins. 
Each method's win ratio is its total points divided by its number of valid comparisons. 
We combine comparisons across all visible/invisible segments and dynamic/static objects.
We then compute Spearman's rank correlation between these win ratios across the 12 methods, weighting each method equally.

As shown in Fig.~\ref{fig:human-study-win-ratio}, Spearman's rank correlations are $0.944$ for \texttt{Perm}, $0.566$ for \texttt{Cont}, and $0.986$ for \texttt{App}, based on 826, 676, and 675 valid comparisons, respectively. These results show strong agreement with human rankings for object permanence and appearance preservation, and moderate agreement for motion continuity. The correlations measure agreement in method rankings, rather than the accuracy of absolute scores.
}

\input{supp_figures/human_study_win_ratio}

%% file: supp_figures/human_study_interface.tex
\begin{figure}[t]
    \centering
    \begin{minipage}[t]{0.35\linewidth}
        \centering
        \vspace{0pt}
        \includegraphics[width=\linewidth]{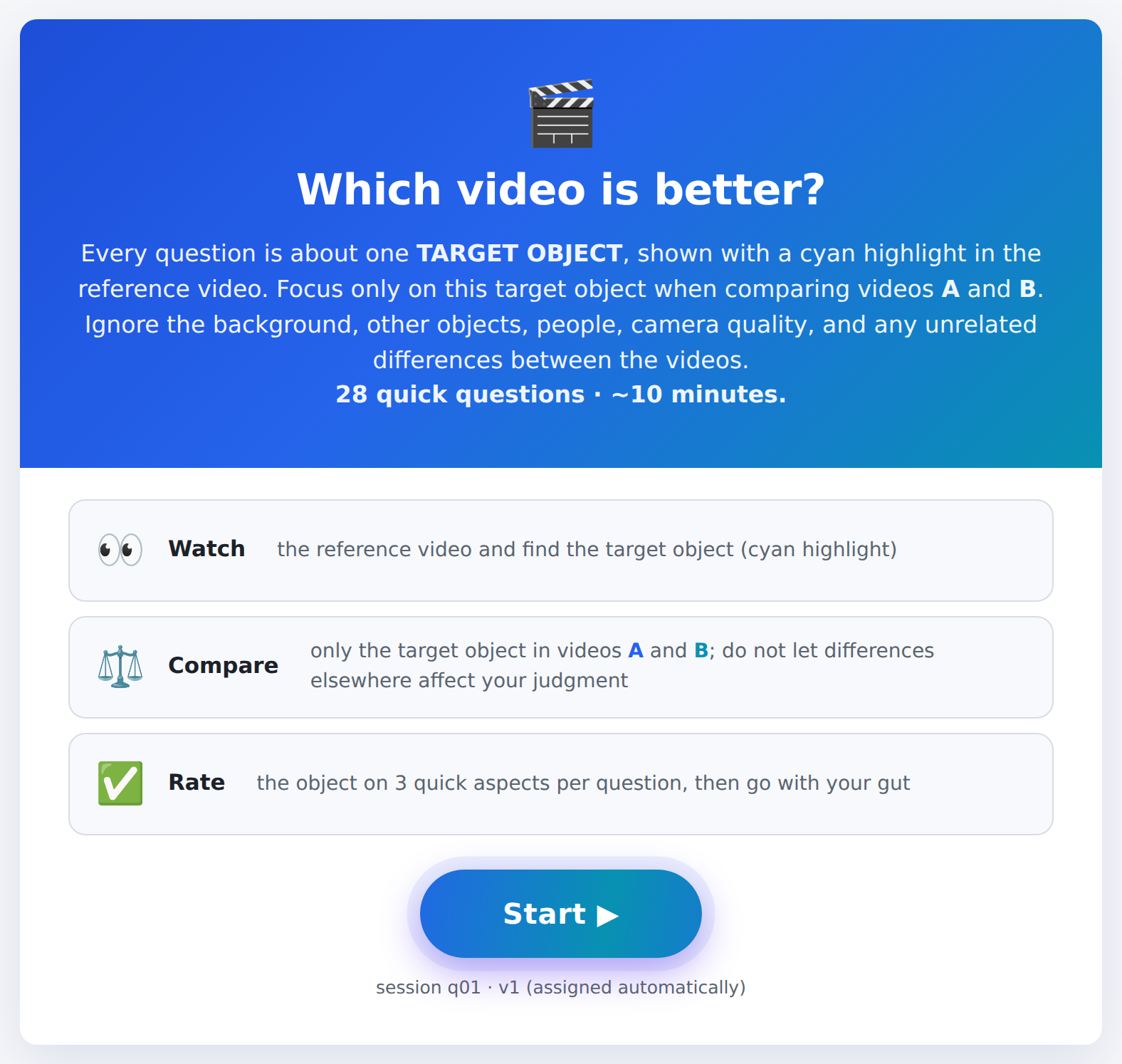}
        \par\smallskip
        {\small\newedit{(a) Instructions}}
    \end{minipage}\hfill
    \begin{minipage}[t]{0.62\linewidth}
        \centering
        \vspace{0pt}
        \includegraphics[width=\linewidth]{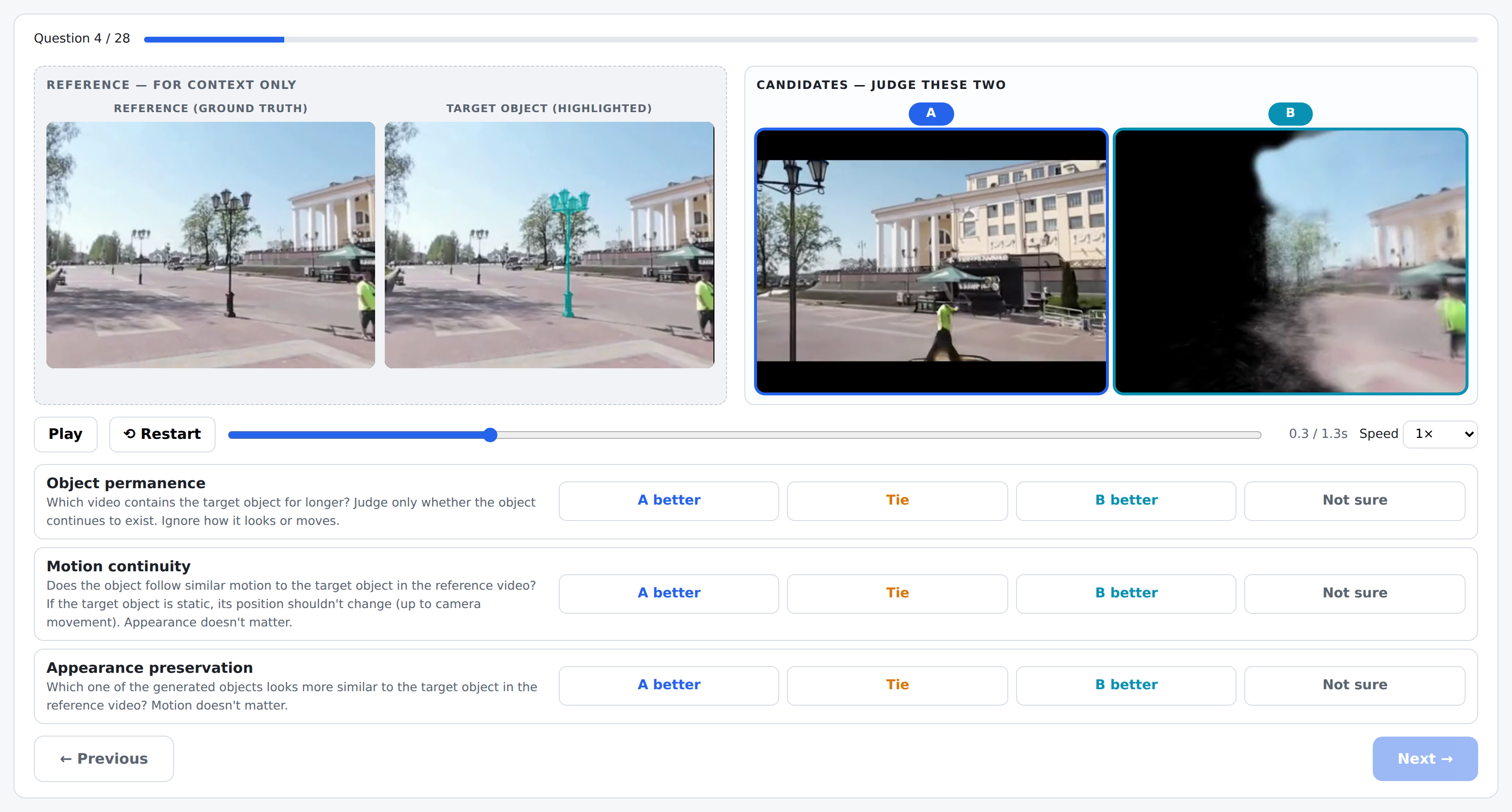}
        \par\smallskip
        {\small\newedit{(b) Example case}}
    \end{minipage}
    \caption{\newedit{\textbf{Human-study interface.} Participant instructions (left) and an example case with a street lamp as the target object (right). Each trial shows the reference video, a copy with the target object highlighted in cyan, and generated videos A and B, along with playback controls and the three metric questions. No participant responses are shown.}}
    \label{fig:human-study-interface}
\end{figure}

%% file: supp_figures/human_study_win_ratio.tex
\begin{figure}[t]
    \centering
    \includegraphics[width=\linewidth]{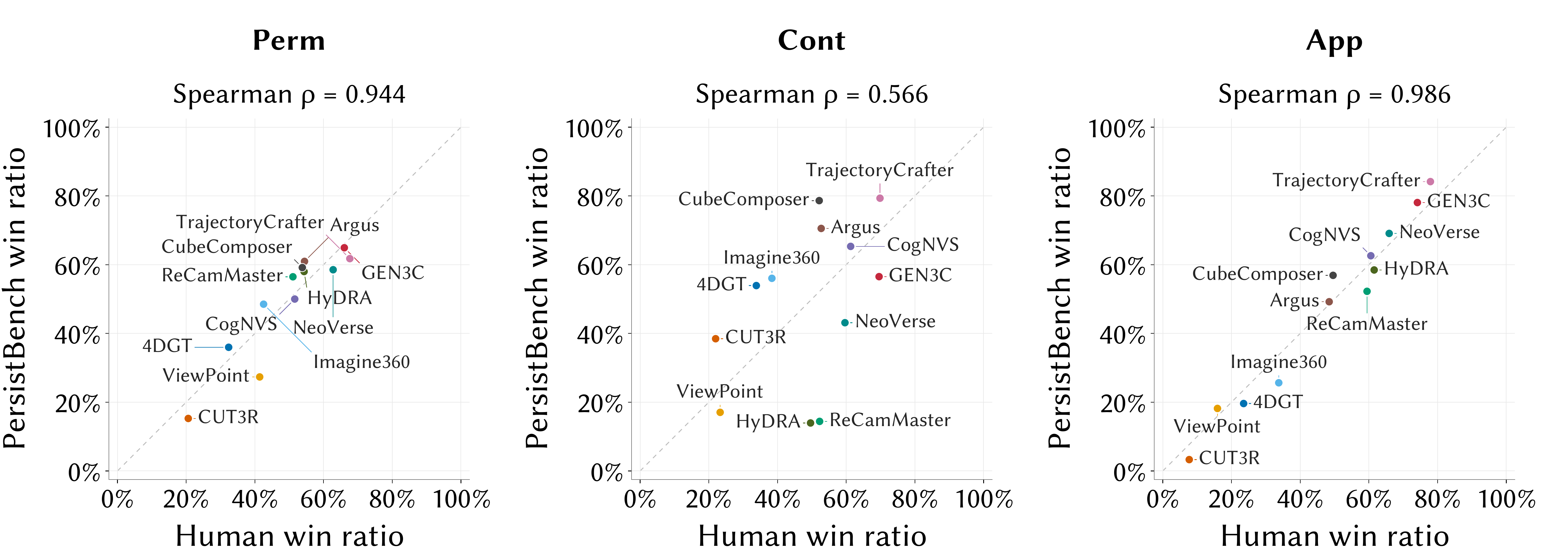}
    \caption{\newedit{\textbf{Alignment with human judgments.} We compare human win ratios ranking (horizontal axis) with \methodname{} win ratios (vertical axis) across the 12 evaluated methods. Each point represents one method. Win ratios assign one point to a win, half a point to a tie, and zero to a loss, using the same valid comparisons across visible/invisible segments and dynamic/static objects. Spearman's correlations are $0.944$ for \texttt{Perm}, $0.566$ for \texttt{Cont}, and $0.986$ for \texttt{App}.}}
    \label{fig:human-study-win-ratio}
\end{figure}

%% file: supp/sections/3_additional_experiments.tex
\newedit{
\section{Additional Experiments}
\label{sec:additional-experiments}

\subsection{Evaluation on Longer Videos}
\label{sec:longer-videos}
To show that our data construction and evaluation pipeline can be easily scaled beyond the current clip lengths, we evaluate 4DGT~\citep{xu20254dgt} on 10 additional dynamic object sequences of approximately 60 seconds each, constructed from longer $360^\circ$ videos. We report \texttt{Perm}, \texttt{Cont}, and \texttt{App} separately for visible and invisible segments together with our default dynamic subset results.

\begin{table}[htbp]
    \centering
    \small
    \setlength{\tabcolsep}{5pt}
    \begin{tabular}{@{}lrrr@{}}
    \toprule
    Duration & \texttt{Perm} & \texttt{Cont} & \texttt{App} \\
    \midrule
    $\sim$10\,s (default) & 3.89\% (96.21\%) & 61.12\% (91.25\%) & 23.96\% (79.98\%) \\
    $\sim$60\,s & 2.69\% (42.07\%) & 45.32\% (67.30\%) & 61.44\% (48.13\%) \\
    \bottomrule
    \end{tabular}
    \vspace{5pt}
    \caption{\newedit{\textbf{4DGT evaluation on longer videos.} Both rows report dynamic-object results, with invisible scores followed by visible scores in parentheses. The longer-video evaluation uses 10 cases with 128 prediction frames per case. Following Tab.~\ref{tab:result-table}, case scores are equally weighted for each metric. The higher invisible-segment \texttt{App} score on longer videos is averaged over tracked frames from only 3 of the 10 cases with valid scores, so it should not be interpreted as improved appearance preservation across the full set.}}
    \label{tab:longer-videos}
\end{table}
}

\newedit{
\subsection{Category-Wise Results}
\label{sec:category-wise-results}
Fig.~\ref{fig:category-wise-results} reports the three memory scores across object categories for all 12 evaluated methods, separately for dynamic and static objects. Performance varies across categories, and a method's strength in one memory property does not always extend to the others. For example, GEN3C achieves high \texttt{App} scores across both dynamic and static categories, while its \texttt{Cont} ranking varies across categories. Within the dynamic subset, several methods obtain lower \texttt{Cont} scores for animals than for decorations. The static subset also shows category-dependent differences in scores and method rankings. These results provide a more detailed view of model performance than the overall averages.
}

\input{supp_figures/category_wise_results}

\newedit{
\subsection{Temporal Stability Analysis}
\label{sec:temporal-stability}

\input{supp_figures/temporal_analysis}

To examine how visual memory scores change over time, we report mean \texttt{Perm}, \texttt{Cont}, and \texttt{App} scores against normalized progress through both visible and invisible segments in Fig.~\ref{fig:temporal-analysis}. Across the 12 evaluated methods, scores generally decrease slightly over time without sudden changes. 
While some methods change ranking, the overall performance trends remain similar across time. This supports our use of average scores to summarize each method's performance, while the curves provide additional detail on how performance changes within each segment. 

}

\newedit{
\subsection{Motion Continuity and Motion Ambiguity}
\label{sec:motion-ambiguity}

As a dynamic object remains out of view longer, its future motion becomes less certain, and multiple trajectories may be plausible. Our \texttt{Cont} metric compares the predicted object position with the reference and does not account for these alternative trajectories. As shown in the invisible-segment results in Fig.~\ref{fig:temporal-analysis}, \texttt{Cont} remains relatively stable for several methods and decreases gradually for others.
These results support the use of reference-based \texttt{Cont} in our setting, although motion ambiguity may become more significant over longer invisible intervals.
}

%% file: supp_figures/category_wise_results.tex
\begin{figure}[t]
    \centering
    \includegraphics[width=\linewidth]{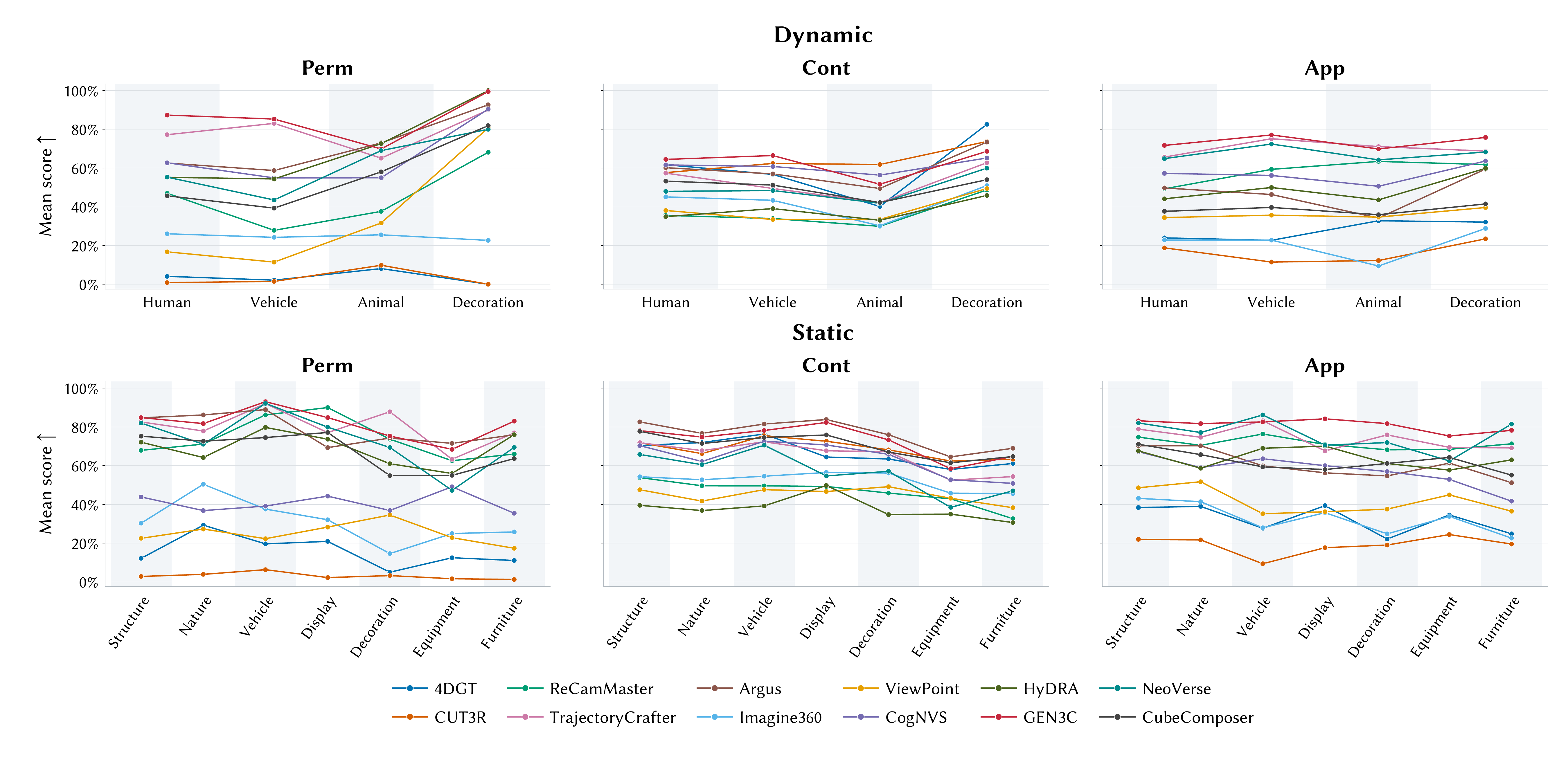}
    \caption{\newedit{\textbf{Category-wise visual memory results.} We report mean \texttt{Perm}, \texttt{Cont}, and \texttt{App} scores (left to right) across the 12 evaluated methods, separately for dynamic objects (top) and static objects (bottom). Each curve represents one method, and higher scores indicate better performance. The results show differences in performance across object categories and memory properties.}}
    \label{fig:category-wise-results}
\end{figure}

%% file: supp_figures/temporal_analysis.tex
\begin{figure}[t]
    \centering
    \includegraphics[width=\linewidth]{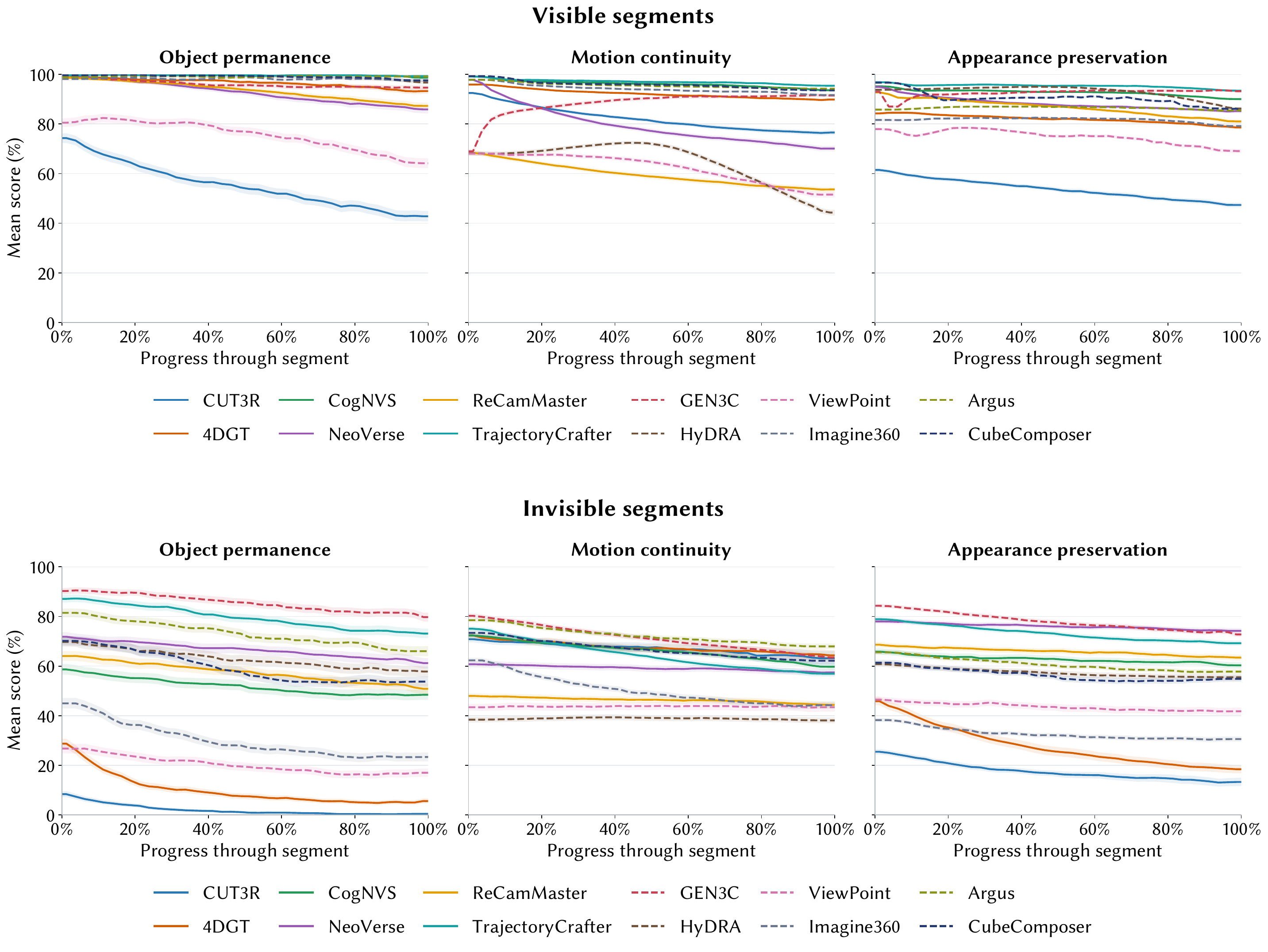}
    \caption{\newedit{\textbf{Temporal analysis of visual memory scores.} Mean \texttt{Perm}, \texttt{Cont}, and \texttt{App} scores (left to right) are plotted against normalized progress through visible segments (top) and invisible segments (bottom) across 12 evaluated methods. 
    Scores generally decrease over time without sudden changes. While some methods change ranking, the overall performance trends remain similar across time, supporting our use of average scores to summarize each method's performance. }}
    \label{fig:temporal-analysis}
\end{figure}

%% file: supp/sections/4_models.tex
\section{Additional Details on Models}
\label{app:models}

\paragraph{Baseline inference protocol.}
For each method, the input is the \texttt{view\_0} perspective video and its camera trajectory, and the query is the target camera trajectory from the paired center view.
We run each baseline once to produce RGB frames for the target view, then evaluate the saved predictions with \methodname{}.
Unless otherwise noted, the benchmark visibility annotations are not provided as model inputs; they are used only for metric computation and for selecting the tracking start frame.
The only exception is HyDRA, whose wrapper uses the visibility mask to split the sequence into a visible conditioning prefix and a post-visible target segment.

\paragraph{Frame staging.}
Different baselines impose different input sizes and temporal lengths.
Before inference, we therefore stage each sequence according to the model interface: for fixed-length models, frames and camera poses are uniformly sampled with rounded \texttt{linspace} indices; for fixed-resolution models, images are center-cropped to the target aspect ratio and resized, while the pinhole intrinsics are updated accordingly.
Models declared with arbitrary length and resolution receive the original frames and camera files directly.
All camera files store OpenCV camera-to-world poses and a pinhole intrinsic matrix.

\begin{table}[t]
\centering
\small
\setlength{\tabcolsep}{5pt}
\begin{tabular}{lcc}
\toprule
Model & Frames & Resolution \\
\midrule
Argus~\citep{argus} & 25 & $384{\times}512$ \\
CogNVS~\citep{cognvs} & 49 & $480{\times}720$ \\
CubeComposer~\citep{cubecomposer} & 27 & -- \\
CUT3R~\citep{cut3r} & -- & -- \\
4DGT~\citep{xu20254dgt} & 128 & $504{\times}672$ \\
GEN3C~\citep{gen3c} & 121 & $704{\times}1280$ \\
HyDRA~\citep{hydra} & 77 & $480{\times}832$ \\
Imagine360~\citep{imagine360} & 32 & $384{\times}512$ \\
NeoVerse~\citep{neoverse} & 81 & $336{\times}560$ \\
ReCamMaster~\citep{recammaster} & 81 & $480{\times}832$ \\
TrajectoryCrafter~\citep{trajcrafter} & 49 & $576{\times}1024$ \\
ViewPoint~\citep{viewpoint} & 49 & $384{\times}512$ \\
\bottomrule
\end{tabular}
\vspace{4pt}
\caption{\textbf{Fixed inference length and resolution for the 12 baselines.} A dash indicates that our evaluator does not impose a fixed value, although the external model may still resize internally.}
\label{tab:app_model_inference}
\end{table}

\subsection{Per-Model Inference Details}
\label{app:per_model_inference}

\paragraph{Argus.}
Argus~\citep{argus} is run through its ground-truth-pose inference path. The model first synthesizes a 360$^\circ$/equirectangular video from the staged perspective input using Stable Video Diffusion image-to-video weights, with source ground-truth poses used to build its motion cache. The final target predictions are obtained by cropping perspective views from the generated equirectangular video according to target rotations and field of view, so target translation is not represented in the final crop. Argus ignores the benchmark visibility masks during inference.

\paragraph{CogNVS.}
CogNVS~\citep{cognvs} estimates depth on the input frames with Depth Anything 3, forward-warps each source frame into the corresponding target camera, and then runs CogVideoX/CogNVS inpainting diffusion over the warped video and masks. It uses an empty prompt, 50 denoising steps, guidance scale 6.0, bf16 inference, a DPM scheduler, model CPU offload, VAE slicing/tiling, and the implementation's fixed random seed of 42. The known cameras are used for the target-view warping stage, while visibility masks are ignored.

\paragraph{CubeComposer.}
CubeComposer~\citep{cubecomposer} projects the perspective input video into equirectangular and cubemap conditioning, completes the panoramic video over temporal windows, and crops target perspective views from the generated panorama. Its cached 3k configuration uses 27 frames, window length 9, cubemap size 768, all six cube faces, diagonal-context conditioning, two history windows, future context, and a global sink token. The method is rotation-centric in this setup: camera translations and visibility masks are ignored during inference.

\paragraph{CUT3R.}
CUT3R~\citep{cut3r} receives the original staged sequence without evaluator-imposed sampling or resizing, although the model internally resizes/crops images for inference. It processes the input sequence recurrently to build a 3D memory, then probes that memory with target ray maps. In our relative setting, target translations are scaled into CUT3R's coordinate system using the ratio between CUT3R-predicted and ground-truth first-to-last input-camera displacement. CUT3R uses its predicted poses, focal lengths, intrinsics, and 3D points, and ignores visibility masks.

\paragraph{4DGT.}
4DGT~\citep{xu20254dgt} converts the benchmark OpenCV camera-to-world poses to the model's OpenGL convention, encodes input frames, cameras, and timestamps into a dynamic Gaussian/video representation, and sequentially renders RGB frames for the requested target cameras. The method is geometry/rendering-centric rather than diffusion-inpainting-centric, and it ignores benchmark visibility masks during inference.

\paragraph{GEN3C.}
GEN3C~\citep{gen3c} uses the staged input frames, cameras, and repeated intrinsics to construct a depth- and camera-conditioned 4D cache. If explicit depth is unavailable, it estimates depth with Depth Anything 3. It then renders target-view warp guidance and synthesizes the final video with the Gen3C-Cosmos-7B video2world pipeline. We use an empty prompt and enable offloading for the diffusion transformer, tokenizer, text encoder, prompt upsampler, and guardrail models. Visibility masks are not used during inference.

\paragraph{HyDRA.}
HyDRA~\citep{hydra} is the only baseline wrapper that uses the benchmark visibility mask during inference. It finds the last visible frame, treats the visible prefix as conditioning, samples both conditioning and target segments to the 77-frame model interface, and restores the full timeline after generation. Target camera-to-world poses are converted into HyDRA camera embeddings, with every fourth camera used by the model. The diffusion run uses Wan2.1 components, the HyDRA checkpoint, 50 denoising steps, classifier-free guidance scale 5.0, and generic positive/negative prompts.

\paragraph{Imagine360.}
Imagine360~\citep{imagine360} converts the perspective input into panorama conditioning, runs the Imagine360/AnimateDiff pipeline with the fixed prompt ``A realistic 360 video.'', and saves generated equirectangular frames. Target predictions are cropped from these ERP frames using relative rotations, with GeoCalib pitch estimates used when enabled by the configuration. As with other panorama-crop baselines, target translation is only weakly represented, and visibility masks are ignored.

\paragraph{NeoVerse.}
NeoVerse~\citep{neoverse} first runs a DA3-based reconstructor to obtain Gaussian splats, rendered intrinsics/extrinsics, and timestamps. It aligns the benchmark cameras to the reconstructor coordinate frame using the first input frame, renders target RGB/depth/mask guidance with the Gaussian renderer, and then completes the result with WanVideoNeoVerse diffusion. The method uses target cameras explicitly through the rendering stage, but ignores visibility masks.

\paragraph{ReCamMaster.}
ReCamMaster~\citep{recammaster} runs a camera-conditioned Wan video diffusion pipeline with the ReCamMaster checkpoint. It captions sampled input frames with its VLM captioner, falling back to a generic prompt if captioning fails, and conditions generation on target camera embeddings relative to the first input camera. The embeddings are downsampled every four frames, and inference uses 50 denoising steps, classifier-free guidance scale 5.0, and seed 0. Visibility masks are ignored.

\paragraph{TrajectoryCrafter.}
TrajectoryCrafter~\citep{trajcrafter} generates a BLIP2 caption when no prompt is supplied, estimates depth with DepthCrafter, and aligns those depths to Depth Anything 3 metric depth using robust scale/shift fitting. It converts source and target OpenCV camera-to-world poses to world-to-camera poses, forward-warps source frames to target views with the metric-aligned depth, and inpaints the warped video and masks with CogVideoX. It uses the first ten input frames as reference and ignores visibility masks.

\paragraph{ViewPoint.}
ViewPoint~\citep{viewpoint} generates a cubemap/equirectangular video from the input with the fixed prompt ``realistic video, natural motion'', converts the generated packed view into a panorama, and crops target perspective views using relative rotations and target intrinsics. It assumes a shared camera center for the crop, so target translations are ignored in this setup. Visibility masks are ignored.

\paragraph{Camera handling.}
The baselines differ substantially in how they use the target camera.
Argus, CubeComposer, Imagine360, and ViewPoint produce an intermediate panoramic or equirectangular video and obtain the final prediction by cropping according to target rotations and intrinsics; target translations are therefore ignored or only weakly represented.
CogNVS, GEN3C, NeoVerse, and TrajectoryCrafter explicitly warp or render target-camera guidance before diffusion completion.
4DGT directly renders target cameras from its reconstructed dynamic representation.
CUT3R stores a recurrent 3D memory from the input and queries it with target ray maps; in our relative setting, target translations are normalized by the ratio between CUT3R-predicted and ground-truth input-camera displacement.
ReCamMaster uses target poses through downsampled relative camera embeddings, taking every fourth target pose.

\paragraph{Evaluation-time auxiliary models.}
The \methodname{} metric is applied after baseline inference and does not re-run the baseline model.
For each predicted video, we resize/copy the ground-truth frames, ground-truth masks, and predictions to the metric resolution of $384{\times}512$.
We use SAM2~\citep{sam2} to track the predicted object from the first visible reference frame, DINOv3~\citep{dinov3} features to localize, DINOv2~\citep{dinov2} to compare object appearance, and the VLM judge described in Sec.~\ref{sec:obj} to verify object presence.
Thus, SAM2, DINOv3, DINOv2, and the VLM judge are evaluation auxiliaries rather than components of the baseline inference pipelines.

%% file: supp/sections/5_visualization.tex
\section{Additional Visualization}
\label{supp:full-results}
We provide additional qualitative results to complement the examples in Fig.~\ref{fig:example-results}.
These visualizations show more static-subset cases across representative baseline predictions, illustrating how the object-centric metrics reflect object permanence, motion continuity, and appearance preservation beyond the main-paper examples.
An interactive visualizer on our \href{https://guangzhaohe.com/persistbench}{project page} allows readers to browse additional benchmark examples and model predictions.

\clearpage
\input{supp_figures/supp_qualitative_dynamic_1}
\clearpage
\input{supp_figures/supp_qualitative_dynamic_2}
\clearpage
\input{supp_figures/supp_qualitative_dynamic_3}
\clearpage
\input{supp_figures/supp_qualitative_static_1}
\clearpage
\input{supp_figures/supp_qualitative_static_2}
\clearpage
\input{supp_figures/supp_qualitative_static_3}
\clearpage

%% file: supp_figures/supp_qualitative_dynamic_1.tex
\begin{figure}[p]
    \centering
    \includegraphics[width=\linewidth,height=0.86\textheight,keepaspectratio]{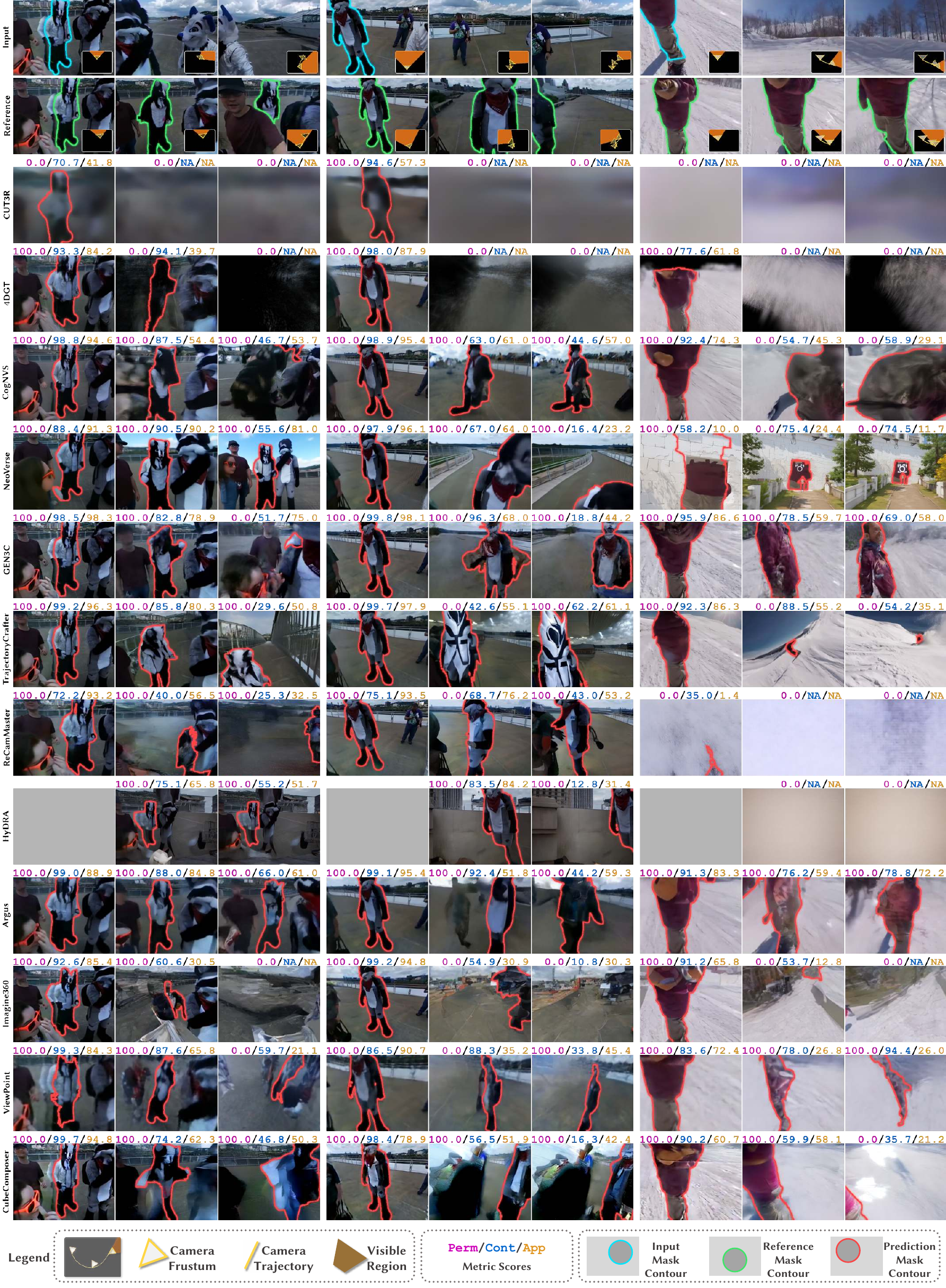}
    \caption{
    \newedit{\textbf{Additional qualitative comparisons on the dynamic subset.}}
    }
    \label{fig:sub-quali-dyn-1}
\end{figure}

%% file: supp_figures/supp_qualitative_dynamic_2.tex
\begin{figure}[p]
    \centering
    \includegraphics[width=\linewidth,height=0.86\textheight,keepaspectratio]{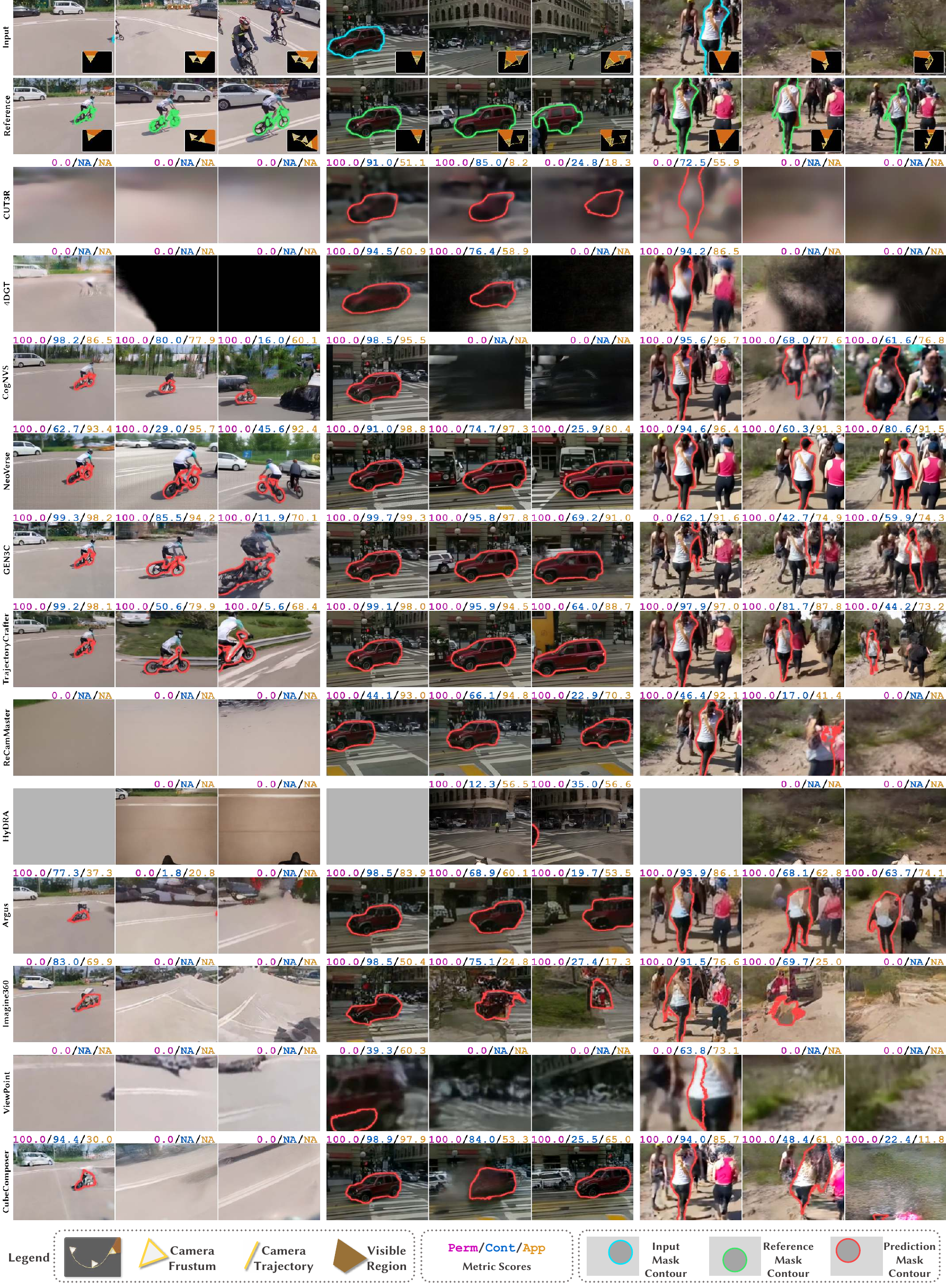}
    \caption{
    \newedit{\textbf{Additional qualitative comparisons on the dynamic subset.}}
    }
    \label{fig:sub-quali-dyn-2}
\end{figure}

%% file: supp_figures/supp_qualitative_dynamic_3.tex
\begin{figure}[p]
    \centering
    \includegraphics[width=\linewidth,height=0.86\textheight,keepaspectratio]{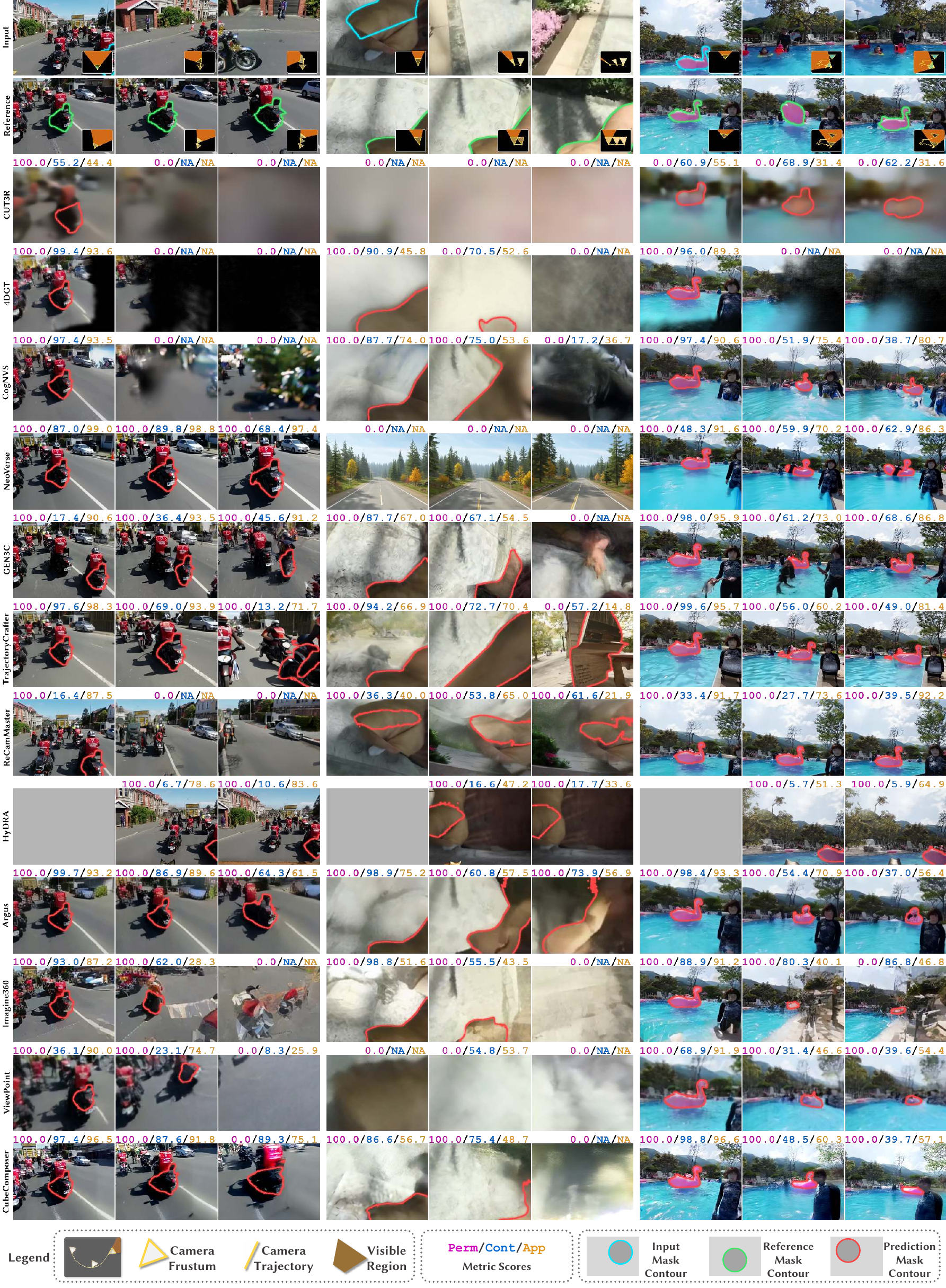}
    \caption{
    \newedit{\textbf{Additional qualitative comparisons on the dynamic subset.}}
    }
    \label{fig:sub-quali-dyn-3}
\end{figure}

%% file: supp_figures/supp_qualitative_static_1.tex
\begin{figure}[p]
    \centering
    \includegraphics[width=\linewidth,height=0.86\textheight,keepaspectratio]{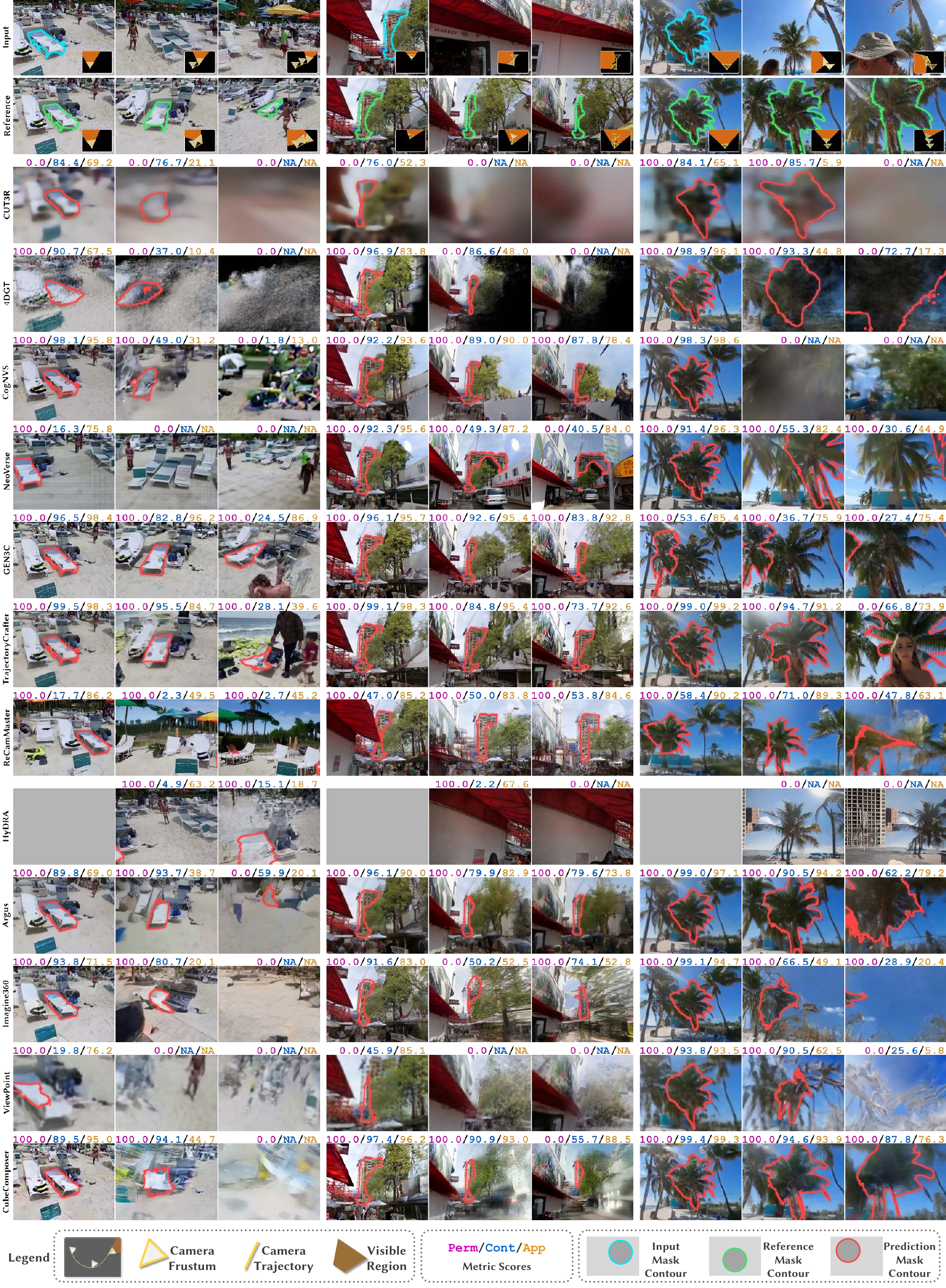}
    \caption{
    \newedit{\textbf{Additional qualitative comparisons on the static subset.}}
    }
    \label{fig:sub-quali-sta-1}
  \end{figure}

%% file: supp_figures/supp_qualitative_static_2.tex
\begin{figure}[p]
    \centering
    \includegraphics[width=\linewidth,height=0.86\textheight,keepaspectratio]{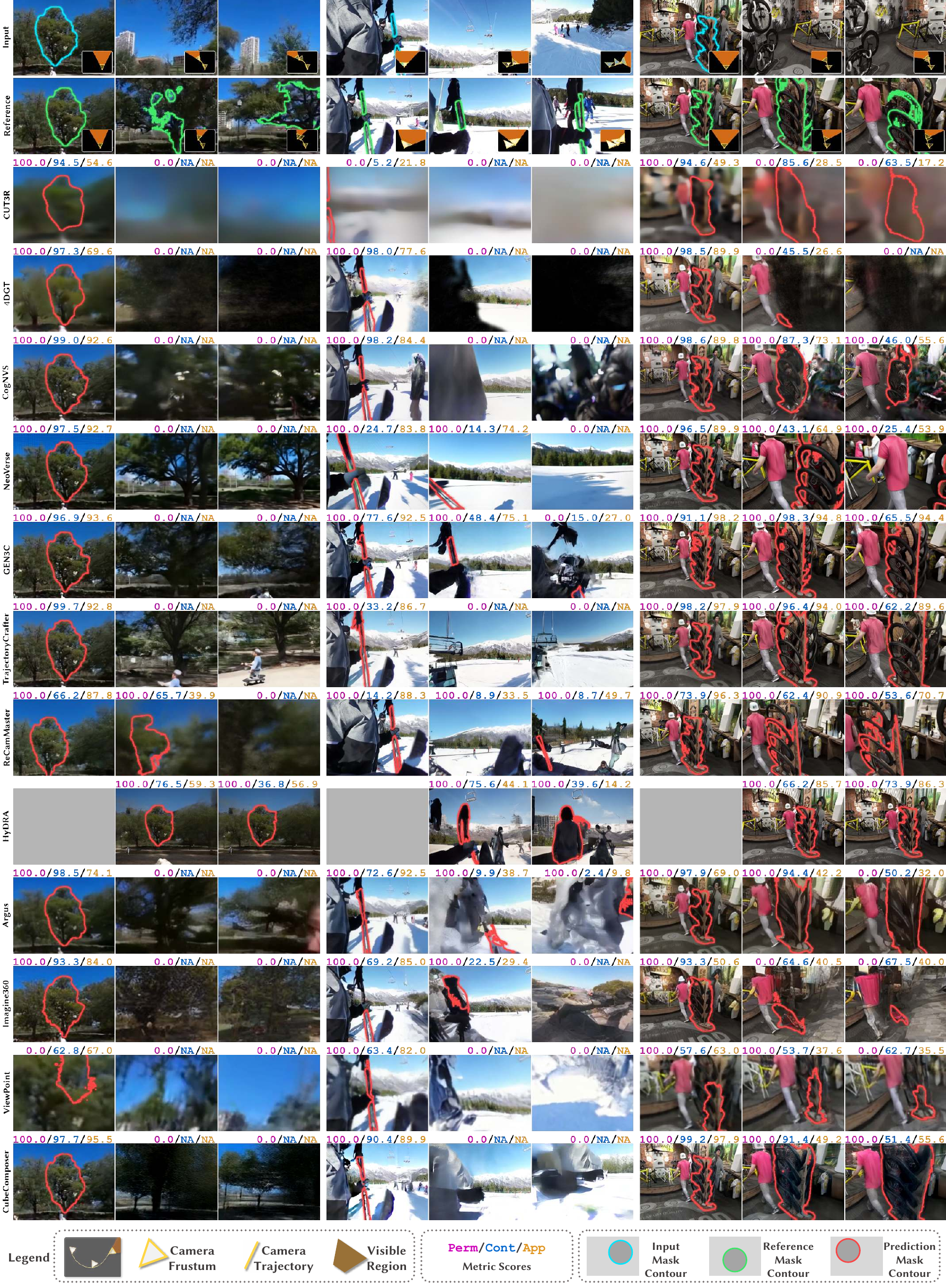}
    \caption{
    \newedit{\textbf{Additional qualitative comparisons on the static subset.}}
    }
    \label{fig:sub-quali-sta-2}
\end{figure}

%% file: supp_figures/supp_qualitative_static_3.tex
\begin{figure}[p]
    \centering
    \includegraphics[width=\linewidth,height=0.86\textheight,keepaspectratio]{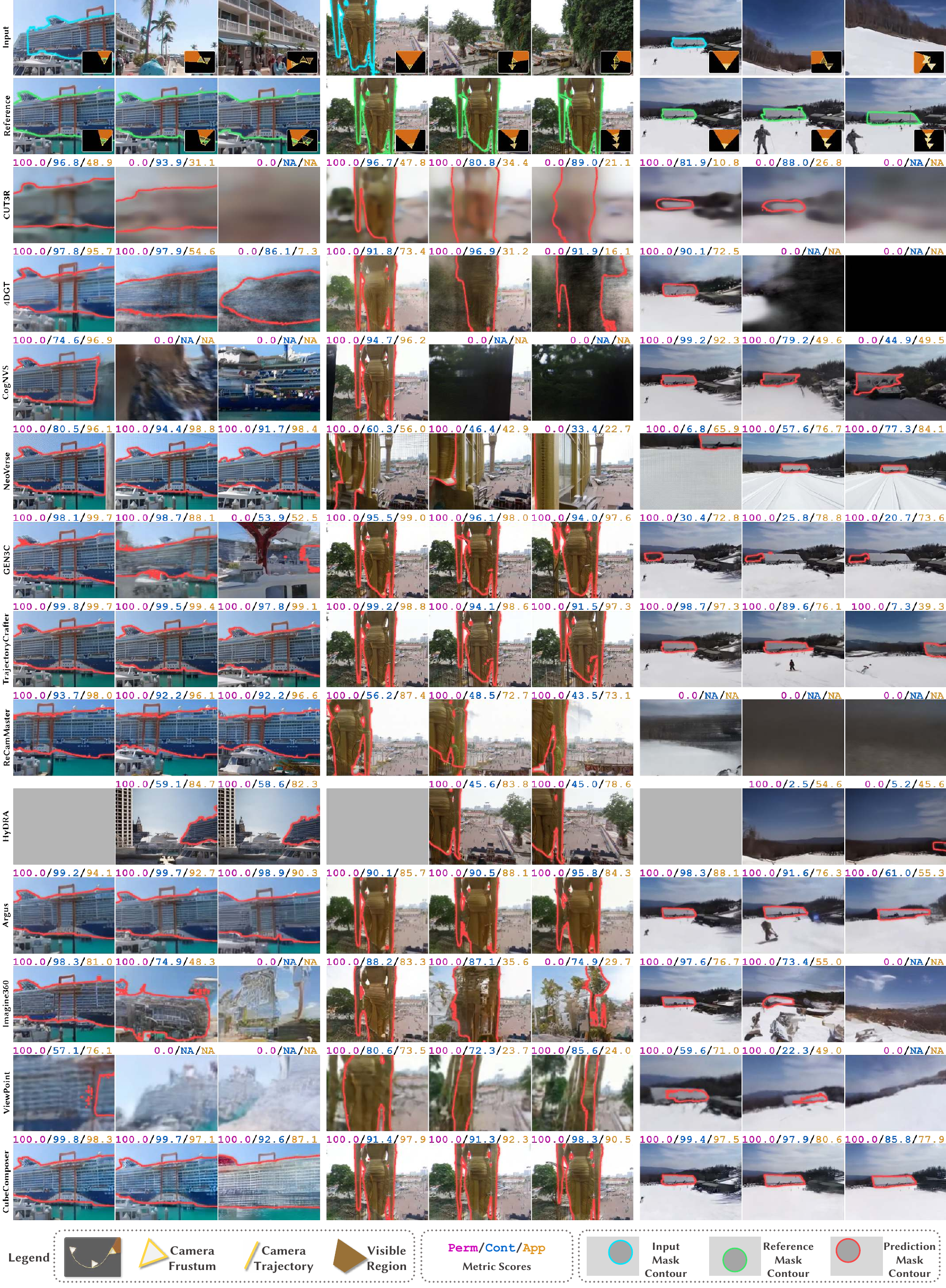}
    \caption{
    \newedit{\textbf{Additional qualitative comparisons on the static subset.}}
    }
    \label{fig:sub-quali-sta-3}
\end{figure}